\documentclass[10pt,twocolumn,letterpaper]{article}

\usepackage[pagenumbers]{cvpr} 

\usepackage{multirow}
\usepackage{pifont}
\usepackage{fontawesome5}
\newcommand{\cmark}{\ding{51}}
\newcommand{\xmark}{\ding{55}}

\usepackage{iftex}
\ifXeTeX
  \usepackage{fontspec}
\fi

\definecolor{cvprblue}{rgb}{0.21,0.49,0.74}
\usepackage[pagebackref,breaklinks,colorlinks,allcolors=cvprblue]{hyperref}

\def\confName{CVPR}
\def\confYear{2027}
\newcommand{\methodName}{Ego2Ego\xspace}

\title{Ego2Ego: Generative Visual Perspective Taking Across Collaborators}

\author{
Chuhan Jiao$^{1}$ \quad
John Gideon$^{2}$ \quad
Anshul Gupta$^{2}$ \quad
Guy Rosman$^{2}$ \quad
Andreas Bulling$^{1}$ \\
$^{1}$University of Stuttgart \qquad
$^{2}$Toyota Research Institute \\[0.4em]
\href{https://chuhan-jiao.github.io/ego2ego/}{\faGlobe\ Project Page}
}

\begin{document}
\twocolumn[{%
\renewcommand\twocolumn[1][]{#1}%
\maketitle
\begin{center}
\centering
\captionsetup{type=figure}
\includegraphics[width=\textwidth]{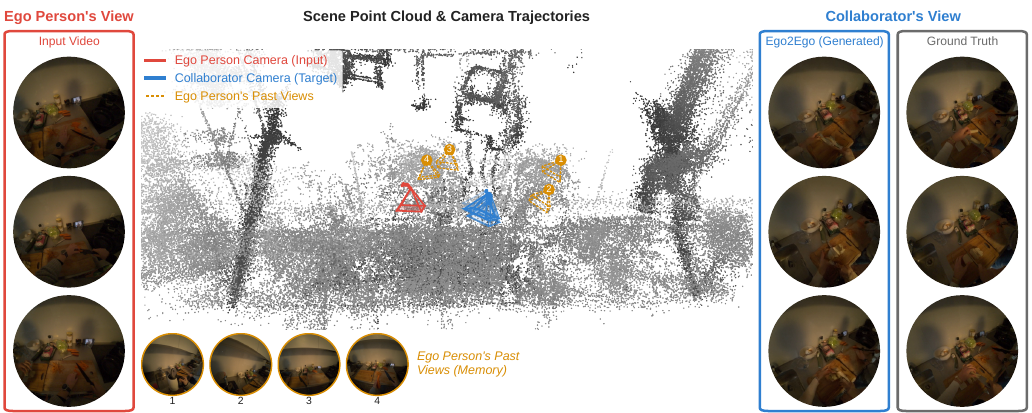}
\caption{\textbf{Generative visual perspective taking.} Given an ego person's egocentric video, the collaborator's camera trajectory, and a scene point cloud, the task is to generate the continuous egocentric video that the collaborator sees at the same moment. Prior perspective-taking studies focused on state estimation, geometric reconstruction, or discrete question answering, while existing cross-view video generation methods either inferred the target view from the input video alone or required an external camera that observes the target person. In contrast, collaborators often face each other or attend to disjoint workspace regions and leave each other's field of view. \methodName provides pretrained video diffusion models with backbone-agnostic geometric modules to generate collaborator views even when the collaborator is not visible to the ego person. As a potential application, \methodName could help a robot infer what its human collaborator sees and turn single-person recordings into synchronised multi-perspective data.}
\label{fig:teaser}
\end{center}

}]
\begin{abstract}
Visual perspective taking, the ability to infer what another person sees from their position, is fundamental to human collaboration.
However, prior works reduced perspective taking to state estimation, geometric reconstruction, or discrete question answering, whereas existing cross-view video generation methods either inferred the target view from the input video alone or required an external camera that observes the target person.
We introduce generative visual perspective taking: Given an ego person's egocentric video, a scene point cloud, and a collaborator's camera trajectory, the task is to generate the continuous egocentric video from the collaborator's perspective.
This task is challenging given extreme viewpoint shifts, dynamic occlusions, and sparse visual overlap.
We address these challenges with \methodName -- a set of backbone-agnostic geometric modules that provide a pretrained video diffusion model with the 3D scene geometry shared by both interactants.
\methodName renders dense, temporally consistent scene coordinate maps into every view as a shared spatial reference, aligns cross-view tokens in 3D ray space through relative geometric attention, and recovers regions outside the ego person's current view from the ego person's visual memory.
Experiments on the CoMind dataset across three video diffusion backbones show that \methodName consistently improves visual quality, Level-2 scene layout accuracy, and collaborator motion, outperforming state-of-the-art cross-view video generation baselines on every backbone.
By generating the collaborator's egocentric video more accurately, \methodName brings visual perspective taking to human-robot collaboration, where it can help infer what a human collaborator sees and does, and can turn single-person recordings into synchronised multi-perspective videos for robot learning.
\end{abstract}
    
\section{Introduction}
\label{sec:intro}

Perspective taking is the mental and emotional ability to step outside one's own frame of reference and understand a situation from another person's point of view.
Its visual form is \textit{visual perspective taking}: The task of determining whether another person can see an object (Level~1) or how the visual scene looks from their position (Level~2)~\cite{flavell1981young,michelon2006two,moll2006level,samson2010seeing}.
Humans rely on this ability during collaboration~\cite{sebanz2006joint}, for example when handing over an object, when referring to an object only visible to the collaborator, or when avoiding obstruction of another person's view or motion.
Consequently, robots collaborating with humans were argued to require similar abilities~\cite{trafton2005enabling,berlin2006perspective,johnson2005perceptual,fischer2016markerless}.
Prior work
reduced visual perspective taking to estimating symbolic or geometric states~\cite{trafton2005enabling,berlin2006perspective,pandey2010mightability}, reconstructing the visual scene from another viewpoint~\cite{johnson2005perceptual,fischer2016markerless}, or answering spatial questions~\cite{lee2025perspective,goral2024seeing}.
We explore the significantly more challenging task of visually synthesising what the collaborator actually sees.

Existing cross-view video generation methods do not address this task:
Novel view synthesis~\cite{van2026anyview,yu2025trajectorycrafter,jeong2025reangle} and camera-controlled video generation~\cite{bai2025recammaster,li2026cameraanything,zhang2025recapture} only re-film an input video along a new camera trajectory and infer unobserved regions from source observations and learned priors, while exo-to-ego methods~\cite{song2026worldwander,kang2026egox,liu2024exo2egov,exo2egosyn2025} require one or more third-person cameras that observe the target person and their workspace.
These assumptions break down in collaborative perspective taking: Two collaborating individuals often face each other or attend to disjoint regions of the workspace, so much of what one person sees is not currently observed by the other, and the collaborator whose view is to be generated is frequently outside the ego person's field of view (see \cref{fig:teaser}).

In this work we study \emph{generative visual perspective taking} between two human collaborators:
Given the posed egocentric video of one person, the \emph{ego person}, with access to earlier frames of the same recording, a scene point cloud, and the camera trajectory of the other person, the \emph{collaborator}, the task is to generate the egocentric video that the collaborator sees at any point in time.
While pretrained video diffusion models provide strong priors over scene appearance and human motion, they cannot directly relate two views separated by large viewpoint changes.
Our key insight is that collaboration takes place in a shared workspace: its static 3D geometry, together with the collaborator's camera trajectory, determines which surfaces the collaborator sees and where, while the ego person's observations provide appearance cues that the video prior can build on for many of these surfaces.
We thus propose \methodName -- a set of backbone-agnostic geometric modules that ground video generation in this shared geometry and visual memory.
First, to handle extreme viewpoint shifts and dynamic occlusions, dense geometric rendering rasterises the point cloud into a scene coordinate map for every view, so that the same 3D point receives the same value across views and serves as a shared spatial reference.
Second, to relate two independently moving cameras, relative geometric attention extends single-trajectory ray-aligned attention~\cite{zhang2026unified} to align tokens between the ego person's and the collaborator's views, pairing token-wise ray-frame transforms with shared temporal encodings.
Third, to handle sparse visual overlap, the ego person's visual memory selects earlier ego-person frames that best cover the collaborator's view, recovering regions outside the current clip.

We evaluate \methodName on CoMind~\cite{gavryushin2026comind} across three backbones: the general-purpose video models Wan2.2-5B and Wan2.1-14B~\cite{wan2025}, and AnyView~\cite{van2026anyview}, which is pretrained for dynamic view synthesis.
Besides visual quality, we evaluate whether the generated video reproduces the scene layout from the collaborator's viewpoint (Level-2 perspective taking) and captures the collaborator's head and hand motions, which matter for learning manipulation policies from human videos~\cite{qin2022dexmv,wang2023mimicplay,kareer2025egomimic}.
\methodName consistently improves all three backbones and outperforms the baselines built on the same backbone on all metrics, including Level-2 scene layout, and also on commonly used video quality metrics when the collaborator's head stays outside the ego person's view.
Notably, \methodName on general-purpose backbones without cross-view pretraining surpasses AnyView on most metrics despite AnyView's large-scale multi-view pretraining, while adding \methodName to AnyView brings further gains.

In summary, our contributions are three-fold:
\begin{itemize}[nosep,leftmargin=*]
\item We introduce the task of \textit{generative visual perspective taking between collaborators}, and evaluate existing novel view synthesis, camera-controlled, and exo-to-ego video generation methods on the CoMind dataset.
\item We propose \methodName, a backbone-agnostic geometric method that combines shared 3D scene coordinates, relative ray-space attention, and the ego person's visual memory.
\item Across three video diffusion backbones, \methodName consistently outperforms the cross-view video generation baselines built on the same backbone in visual quality, Level-2 scene layout, and collaborator motion.
\end{itemize}

Because its inputs require only a wearable camera, a one-time scene scan, and a camera trajectory, \methodName is promising for human-robot collaboration in two complementary ways:
When the ego person is a robot collaborating with a human, visual perspective taking could allow the robot to infer what its human collaborator sees from their head trajectory, helping the robot plan complementary next actions and avoid redundant work.
Conversely, when training a robot policy to act as the collaborator, \methodName can turn single-person recordings into synchronised multi-perspective videos, which are costly to capture~\cite{wiederhold2023hoh,gavryushin2026comind} but valuable for visuomotor policy learning~\cite{chi2023diffusionpolicy,kareer2025egomimic}.

\section{Related Work}
\label{sec:rw}

\paragraph{Visual perspective taking.}
Visual perspective taking, the ability to infer whether another person can see an object (Level~1) and how the scene appears from their viewpoint (Level~2)~\cite{flavell1981young,michelon2006two,moll2006level,samson2010seeing}, is central to spatial cognition and human collaboration~\cite{sebanz2006joint}.
In human-robot interaction, perspective taking has been modelled through symbolic and geometric reasoning, enabling robots to resolve ambiguous spatial references~\cite{trafton2005enabling}, interpret demonstrations from the teacher's point of view~\cite{berlin2006perspective}, reason about what a human can see and reach~\cite{pandey2010mightability}, and reconstruct the scene from the human's viewpoint for action recognition and visibility estimation~\cite{johnson2005perceptual,fischer2016markerless}.
Recent benchmarks show that vision-language models struggle with visual perspective taking~\cite{goral2024seeing} and 3D spatial reasoning~\cite{yang2025thinking,ma20253dsrbench}, though spatial QA data~\cite{chen2024spatialvlm} and coordinate transforms or mental imagery~\cite{lee2025perspective} improve accuracy.
However, these works represent another person's perspective through state estimation, geometric reconstruction, or discrete question answering.
In contrast, we formulate visual perspective taking as a conditional video generation problem and synthesise the continuous egocentric video that a collaborator sees during a shared task.

\paragraph{Cross-view and egocentric video generation.}
Cross-view video generation synthesises a video from a target viewpoint that differs from the source view.
Camera-controlled generation conditions a video model on a camera trajectory, encoded either from camera extrinsics~\cite{wang2024motionctrl,bai2025recammaster} or as pixel-wise Plücker ray maps~\cite{sitzmann2021lfn,he2025cameractrl,bahmani2025vd3d,bahmani2025ac3d,xu2024camco}, and has been applied to exploring dynamic scenes~\cite{he2025cameractrl2}, generating synchronised multi-camera videos~\cite{bai2025syncammaster}, and re-filming an input video along a new trajectory~\cite{bai2025recammaster,li2026cameraanything,zhang2025recapture}.
Novel view synthesis (NVS) warps the input views through estimated geometry before inpainting~\cite{yu2024viewcrafter,yu2025trajectorycrafter,ren2025gen3c,jeong2025reangle}, or learns view transformations from multi-view data for dynamic scenes~\cite{van2024generative,wu2025cat4d,van2026anyview} and objects~\cite{xie2025sv4d}.
Exocentric-to-egocentric (exo-to-ego) methods translate third-person observations into the actor's first-person view from a single exocentric image~\cite{egoworld2025}, a monocular exocentric video~\cite{luo2024put,kang2026egox,song2026worldwander,groundedexo2ego2026}, or multiple exocentric cameras~\cite{liu2024exo2egov,exo2egosyn2025}.
In addition, E3C~\cite{gu2026e3c} generates an egocentric video from an environmental point-cloud memory controlled by the wearer's 3D body motion, and concurrent work ME-World~\cite{chung2026meworld} generates multi-agent egocentric views from each agent's initial frame, head trajectory, and body pose.
Overall, these methods assume either that the target view can be inferred from the input video alone (video re-filming and NVS), that external cameras observe the target person (exo-to-ego), or that the target person's own initial frame and body pose are given (E3C and ME-World).
In contrast, ego-to-ego perspective taking has neither an external camera nor the collaborator's image or body pose: given only the ego person's video, a scene point cloud, and the collaborator's trajectory, the model must synthesise the scene layout and collaborator motion from a largely non-overlapping view.

\paragraph{Geometry-aware attention and visual memory.}
Dense 3D coordinates are predicted from images for camera relocalisation~\cite{shotton2013scene,brachmann2017dsac,brachmann2023ace}, object pose estimation~\cite{wang2019normalized}, and pointmap-based 3D reconstruction~\cite{wang2024dust3r}, while multi-view transformers inject camera geometry into attention through camera-level $SE(3)$ transforms~\cite{miyato2024gta}, pinhole projection matrices~\cite{li2025prope}, token-wise ray-aligned frames~\cite{zhang2026unified}, or depth-dependent ray encodings~\cite{wu2026rayrope}.
Meanwhile, long-horizon video world models condition on past frames~\cite{song2025historyguided}, retrieve them from the camera's own trajectory by field-of-view overlap~\cite{xiao2026worldmem,yu2025context} or surfel-indexed visibility~\cite{li2025vmem}, or maintain a pointmap-based spatial memory~\cite{wu2025spatialmemory}.
However, token-wise ray-aligned attention~\cite{zhang2026unified} encodes only a single camera trajectory with 2D spatial encodings, while existing memory mechanisms retrieve a camera's own past rather than another person's observations.
In contrast, \methodName renders dense 3D scene coordinate maps to supply surface depth, extends ray-aligned attention with shared temporal rotary encodings across the ego person's and the collaborator's views, and selects past ego-person frames via maximum coverage~\cite{nemhauser1978analysis}.

\section{Method}
\label{sec:method}

\begin{figure*}[t]
\centering
\includegraphics[width=0.92\textwidth]{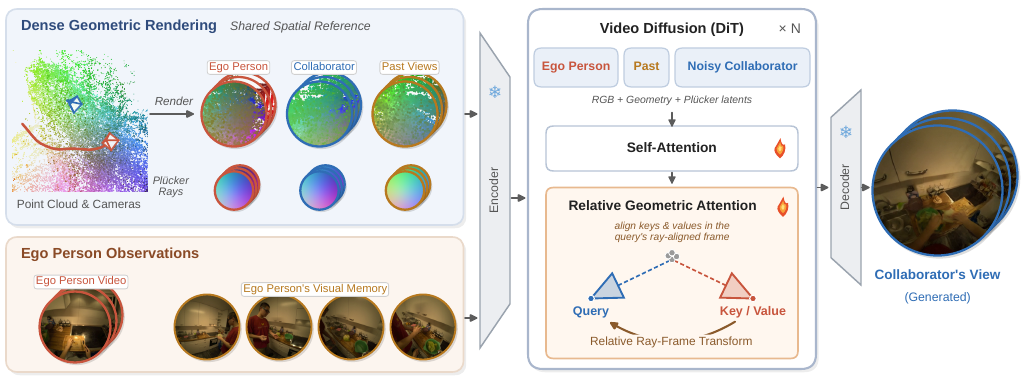}
\caption{\textbf{Overview of \methodName.} Given the ego person's egocentric video, a scene point cloud, and both camera trajectories, \methodName generates the collaborator's egocentric video with a pretrained video diffusion transformer equipped with three geometry-grounded modules. All conditions are encoded by the frozen VAE and form a single token sequence with the noisy collaborator latent. (1)~Since the same surface appears entirely different from the two viewpoints and hands occlude the scene, \emph{dense geometric rendering} rasterises the point cloud into every view, colouring each pixel by its 3D coordinate in a shared frame, so that the same scene point carries the same colour across views and provides a shared spatial reference. (2)~To relate tokens across substantially different viewpoints, \emph{relative geometric attention} aligns the geometric components of keys and values to each query token's ray-aligned frame. (3)~To recover regions the ego person does not currently observe, the ego person's \emph{visual memory} supplies past ego-person frames together with their renderings.}
\label{fig:method}
\end{figure*}

Visual perspective taking requires answering two questions about the collaborator~\cite{flavell1981young}: whether they can see an object (Level~1), and how the visual scene appears from their perspective (Level~2).
Our key insight is that the static 3D scene geometry shared by both collaborators determines which surfaces the collaborator can see and where they project in the other person's view, so that only their colour and texture, together with the dynamic hands and bodies, remain to be transferred from the ego person's observations or synthesised by a video prior.
\methodName builds on a pretrained latent video diffusion transformer~\cite{peebles2023dit} (\cref{sec:backbone}) and introduces three backbone-agnostic geometric modules (see \cref{fig:method}): dense geometric rendering at the input (\cref{sec:dgr}), relative geometric attention within attention (\cref{sec:rga}), and the ego person's visual memory as historical context (\cref{sec:memory}; details in the supplementary material).

\subsection{Problem Formulation}
\label{sec:formulation}

Given the egocentric video of one person, the \emph{ego person}, during a collaborative task, the goal is to visually generate the simultaneous egocentric video of the other person, the \emph{collaborator}, along their camera trajectory.
Formally, let $\mathbf{V}^{c}=\{I^{c}_{l}\}_{l=1}^{L}$ denote the collaborator's target video of $L$ frames and $\Phi^{c}=\{T^{c}_{l}\}_{l=1}^{L}$ their camera trajectory, where $T^{c}_{l}\in SE(3)$ is the camera-to-world pose at frame $l$.
Let $\mathbf{V}^{p}=\{I^{p}_{l}\}_{l\le L}$ and $\Phi^{p}=\{T^{p}_{l}\}_{l\le L}$ denote the ego person's egocentric video and camera trajectory, where frames $l\in\{1,\dots,L\}$ are synchronised with the target clip and frames $l\le0$ come from the ego person's preceding recording in the same session.
Given the camera intrinsics $K^{p},K^{c}$ and a static scene point cloud $\mathcal{P}=\{\mathbf{x}_{n}\in\mathbb{R}^{3}\}_{n=1}^{N}$, expressed in the collaborator's first camera frame, we model $p(\mathbf{V}^{c}\mid\mathcal{C})$ with $\mathcal{C}=\{\mathbf{V}^{p},\Phi^{p},\Phi^{c},K^{p},K^{c},\mathcal{P}\}$.

\subsection{Base Model}
\label{sec:backbone}

Most large-scale egocentric datasets record a single camera wearer~\cite{grauman2022ego4d,grauman2024ego,wang2023holoassist}, while datasets that capture multiple synchronised egocentric views~\cite{gavryushin2026comind} are much smaller.
We therefore use a pretrained latent video diffusion transformer (DiT) and its priors over scene appearance and human motion.
A frozen VAE encodes the ego person's current clip $\mathbf{V}^{p}_{1:L}$ and the collaborator's target clip $\mathbf{V}^{c}$ into latents $\mathbf{z}^{p}$ and $\mathbf{z}^{c}_{0}$.
At diffusion time $t\in[0,1]$, clean ego-person tokens $\mathbf{z}^{p}$ and noisy collaborator tokens $\mathbf{z}^{c}_{t}=(1-t)\,\mathbf{z}^{c}_{0}+t\,\boldsymbol{\epsilon}$ (with $\boldsymbol{\epsilon}\sim\mathcal{N}(\mathbf{0},\mathbf{I})$) are concatenated into a single token sequence and processed jointly by self-attention.
Following AnyView~\cite{van2026anyview}, each view is conditioned on its pixel-wise Plücker ray map~\cite{sitzmann2021lfn}, which is encoded by the frozen VAE and concatenated with the view's RGB latents along the channel dimension.
We train the model with the rectified-flow velocity objective~\cite{liu2023rectified,lipman2023flow} on the collaborator tokens,
\begin{equation}
\mathcal{L}_{\text{RF}}=\mathbb{E}_{\mathbf{z}^{c}_{0},\boldsymbol{\epsilon},t}\big\lVert\mathbf{v}_{\theta}(\mathbf{z}^{c}_{t},t,\mathcal{C})-(\boldsymbol{\epsilon}-\mathbf{z}^{c}_{0})\big\rVert_{2}^{2},
\label{eq:rf}
\end{equation}
combined with a latent motion consistency loss $\mathcal{L}_{\text{mot}}$~\cite{jiang2026lamo} with weight $\lambda$ supervising global frame-to-frame latent motion ($\mathcal{L}=\mathcal{L}_{\text{RF}}+\lambda\,\mathcal{L}_{\text{mot}}$).

\subsection{Dense Geometric Rendering}
\label{sec:dgr}

Plücker ray maps describe camera rays as lines in free space rather than where they intersect a surface, and depth maps rendered from $\mathcal{P}$ do not expose cross-view correspondences because depth is measured relative to each camera.
To determine surface visibility and identify which 3D point each pixel observes across views, we introduce \emph{dense geometric rendering} (DGR).
Whereas scene coordinate regression~\cite{shotton2013scene,brachmann2017dsac,brachmann2023ace} and NOCS~\cite{wang2019normalized} predict canonical 3D coordinates from images for pose estimation, we invert this direction: We rasterise $\mathcal{P}$ with depth buffering into each camera view $v$ (ego person and collaborator) at frame $l$, assigning each pixel $\mathbf{u}$ the 3D coordinate $\mathbf{x}^{v}_{l}(\mathbf{u})$ of its nearest visible point in the shared frame:
\begin{equation}
G^{v}_{l}(\mathbf{u})=\tfrac{1}{2}+\tfrac{1}{2}\,\mathrm{clip}\big(\mathbf{x}^{v}_{l}(\mathbf{u})/s,\,-1,\,1\big),
\label{eq:dgr}
\end{equation}
where $s$ is a metric scene scale in metres.

DGR serves two roles in visual perspective taking:
First, because $\mathbf{x}^{v}_{l}(\mathbf{u})$ is expressed in a shared coordinate frame, pixels observing the same static 3D point in any view or frame receive the same value, reducing cross-view correspondence to matching coordinates.
Second, depth-buffered rasterisation determines which static surfaces are visible from the collaborator's position (Level~1) and where they project in the collaborator's view, providing the geometric layout for Level-2 perspective taking.
Because $G^{v}_{l}$ is rendered from the static scene, it stays intact under hand and body occlusions and defines the 3D layout even outside the ego person's view.
The frozen VAE encodes $G^{v}\in[0,1]^{3}$, and the resulting latent is concatenated with the RGB and Plücker latents.

\subsection{Relative Geometric Attention}
\label{sec:rga}

DGR gives only the uncoloured static layout, so the observed appearance of static surfaces and moving objects must come from ego-person tokens through cross-view attention.
However, standard DiT positional encodings only index token positions within each video and do not tell attention how two independently moving cameras relate in 3D, as Level-2 perspective taking~\cite{michelon2006two} requires.
UCPE~\cite{zhang2026unified} refines the camera-level $SE(3)$ attention transforms of GTA~\cite{miyato2024gta} into token-wise ray-aligned frames, which are world-frame invariant and work for any camera model without depth estimation.
Yet UCPE encodes only one camera trajectory and pairs its ray transform with image-space RoPE~\cite{su2024roformer} instead of cross-view temporal alignment.

To address these limitations, we introduce \emph{relative geometric attention} (RGA), which extends ray-aligned attention to cross-person perspective taking and works together with DGR, whose scene coordinate maps supply the surface depth that ray frames alone leave unspecified.
Specifically, we construct a \emph{ray-aligned coordinate frame} $\Gamma_{i}\in SE(3)$ in the shared 3D frame for every token $i$ across both the ego person's and the collaborator's views, centred at the camera centre with its $z$-axis along the viewing ray through the token centre.
In a parallel self-attention branch, RGA transforms the key $\mathbf{k}_{j}$ and value $\mathbf{v}_{j}$ of token $j$ into the ray-aligned frame of query token $i$:
\begin{equation}
\scalebox{0.86}{$\displaystyle a_{ij}=\mathrm{softmax}_{j}\Big(\tfrac{1}{\sqrt{d}}\,\mathbf{q}_{i}^{\top}\Gamma_{i}\Gamma_{j}^{-1}\mathbf{k}_{j}\Big),\quad
\mathbf{o}_{i}=\textstyle\sum_{j}a_{ij}\,\Gamma_{i}\Gamma_{j}^{-1}\mathbf{v}_{j},$}
\label{eq:rga}
\end{equation}
where $d$ is the head dimension.
Within each head, half of the channels apply $\Gamma_{i}\Gamma_{j}^{-1}\in\mathbb{R}^{4\times4}$ to groups of four channels treated as homogeneous vectors, while the other half apply shared temporal RoPE that assigns identical temporal indices ($l\in\{1,\dots,L\}$) to synchronised ego-person and collaborator frames.

\subsection{Ego Person's Visual Memory}
\label{sec:memory}

Together, DGR and RGA specify the 3D layout of the collaborator's view and relate features across viewpoints.
However, conditioning only on the current clip $\mathbf{V}^{p}_{1:L}$ provides no colour and texture for surfaces outside the ego person's current view.
Because both collaborators move within the same workspace, the ego person has often observed these surfaces earlier, often without current-clip occlusions.
We therefore add $M$ frames from $\mathbf{V}^{p}_{\le0}$ as visual memory.
Since self-attention is quadratic in token count, $M$ must be small, so the selected frames should cover as much of the collaborator's view as possible.
We select these frames by solving a \emph{maximum coverage problem} over the scene point cloud.
Let $\mathcal{L}$ be the set of latent-aligned frame indices of the clip, $\mathcal{P}^{c}_{l}\subseteq\mathcal{P}$ be the points visible to the collaborator at frame $l\in\mathcal{L}$, and $\mathcal{P}_{l'}\subseteq\mathcal{P}$ be those visible in past ego-person frame $l'\le0$.
We select the set $S$ of $M$ past frames that maximises the normalised coverage
\begin{equation}
f(S)=\sum_{l\in\mathcal{L}}\frac{\big|\mathcal{P}^{c}_{l}\cap\bigcup_{l'\in S}\mathcal{P}_{l'}\big|}{|\mathcal{P}^{c}_{l}|}.
\label{eq:coverage}
\end{equation}
Since $f$ is monotone submodular, we maximise it greedily, which achieves at least $1-1/e$ of the optimal coverage~\cite{nemhauser1978analysis}.
Each selected frame is encoded as clean context tokens (\emph{Past} in \cref{fig:method}) with its own DGR map, ray-aligned frames, and negative temporal indices in RGA, so that collaborator tokens retrieve its colour and texture via the same attention as the current clip.

\section{Experiments}
\label{sec:exp}

\subsection{Experimental Setup}
\label{sec:setup}

\begin{table*}[t]
\centering
\small
\setlength{\tabcolsep}{3.2pt}
\resizebox{\textwidth}{!}{%
\begin{tabular}{@{}lllccccccccccc@{}}
\toprule
\textbf{Method} & \textbf{Task} & \textbf{Backbone} & \textbf{CV} & \textbf{PSNR}$\uparrow$ & \textbf{SSIM}$\uparrow$ & \textbf{LPIPS}$\downarrow$ & \textbf{CLIP-I}$\uparrow$ & \textbf{ViCLIP}$\uparrow$ & \textbf{FVD}$\downarrow$ & \textbf{Transl}$\downarrow$ & \textbf{Rot}$\downarrow$ & \textbf{Layout}$\uparrow$ & \textbf{Hand}$\uparrow$ \\
\midrule
\multirow{1}{*}{TrajectoryCrafter$^{\dagger}$~\cite{yu2025trajectorycrafter}} & \multirow{1}{*}{NVS} & \multirow{1}{*}{CogVideoX-5B} & \cmark & 13.13 & 0.312 & 0.642 & 0.72 & 0.50 & 915.1 & 0.48 & 23.5 & 2.1 & 17.6 \\
\cmidrule{1-14}
\multirow{2}{*}{ReCamMaster~\cite{bai2025recammaster}} & \multirow{2}{*}{Camera} & \multirow{2}{*}{Wan2.1-1.3B} & \xmark & 15.64 & 0.435 & 0.512 & 0.82 & 0.84 & 347.8 & 0.40 & 16.2 & 8.2 & 14.6 \\
 & &  & \cmark & 16.68 & 0.488 & 0.483 & 0.83 & 0.86 & 238.3 & 0.32 & 12.1 & 11.4 & 22.1 \\
\cmidrule{1-14}
\multirow{1}{*}{CameraAnything$^{\ddagger}$~\cite{li2026cameraanything}} & \multirow{1}{*}{Camera} & \multirow{1}{*}{Wan2.1-14B} & \xmark & 16.70 & 0.479 & 0.461 & 0.84 & 0.88 & 221.8 & \underline{0.30} & 10.5 & 15.7 & 20.2 \\
\cmidrule{1-14}
\multirow{2}{*}{WorldWander~\cite{song2026worldwander}} & \multirow{2}{*}{Exo-to-ego} & \multirow{2}{*}{Wan2.2-5B} & \xmark & 15.94 & 0.449 & 0.516 & 0.81 & 0.82 & 288.7 & 0.48 & 25.0 & 6.1 & 17.5 \\
 & &  & \cmark & 16.07 & 0.451 & 0.505 & 0.82 & 0.86 & 239.8 & 0.45 & 21.1 & 8.7 & 15.8 \\
\cmidrule{1-14}
\multirow{2}{*}{EgoX~\cite{kang2026egox}} & \multirow{2}{*}{Exo-to-ego} & \multirow{2}{*}{Wan2.1-14B} & \xmark & 17.70 & 0.525 & 0.447 & 0.83 & 0.84 & 354.2 & 0.32 & 13.3 & 13.5 & 21.1 \\
 & &  & \cmark & 17.56 & 0.514 & 0.456 & 0.83 & 0.87 & 281.6 & 0.33 & 12.1 & 14.7 & 21.2 \\
\cmidrule{1-14}
\multirow{2}{*}{AnyView~\cite{van2026anyview}} & \multirow{2}{*}{NVS} & \multirow{2}{*}{Cosmos-2B} & \xmark & 16.32 & 0.465 & 0.561 & 0.79 & 0.67 & 1196.4 & 0.59 & 18.7 & 1.1 & 17.4 \\
 & &  & \cmark & 18.23 & 0.513 & 0.400 & \underline{0.87} & 0.89 & 228.0 & 0.43 & 9.3 & 15.9 & \underline{27.1} \\
\midrule
\multirow{3}{*}{\textbf{\methodName (ours)}} & \multirow{3}{*}{Ego-to-ego} & Wan2.2-5B & \xmark & 18.57 & 0.538 & 0.356 & 0.84 & 0.88 & 181.2 & 0.32 & \underline{9.0} & 20.7 & 24.9 \\
 &  & Wan2.1-14B & \xmark & \underline{18.89} & \underline{0.551} & \underline{0.343} & 0.85 & \underline{0.90} & \textbf{164.7} & \textbf{0.28} & 10.1 & \textbf{26.1} & 26.9 \\
 &  & AnyView & \cmark & \textbf{19.57} & \textbf{0.556} & \textbf{0.306} & \textbf{0.89} & \textbf{0.92} & \underline{177.1} & 0.41 & \textbf{8.8} & \underline{23.6} & \textbf{30.9} \\
\bottomrule
\end{tabular}}
\caption{\textbf{Comparison to cross-view video generation methods on CoMind.} CV: initialised from the cross-view pretraining released by the authors. Layout and Hand: Layout mIoU and Hand-mIoU. Bold: best; underlined: second best. $^{\dagger}$Released model, not fine-tuned. $^{\ddagger}$Trained only without cross-view pretraining (\cref{sec:setup}).}
\label{tab:main}
\end{table*}

\noindent\textbf{Dataset.}
We evaluated our method on CoMind~\cite{gavryushin2026comind}, which captures pairs of participants collaborating on everyday tasks while wearing egocentric glasses that record synchronised fisheye videos and camera poses together with a scene point cloud per environment.
We extracted 49-frame clips at 10\,fps and $256\times256$ resolution in both directions (using either person as the ego person), obtaining 7{,}670 training and 1{,}020 test clips from disjoint sessions. All test clips were from scenes not shown in training.
In 46\% of test clips, the collaborator's head never entered the ego person's view, and it was visible in at least half of the frames in only 20\%.

\noindent\textbf{Baselines.}
We compared with representative cross-view video generation methods from three families: novel view synthesis (TrajectoryCrafter~\cite{yu2025trajectorycrafter}, AnyView~\cite{van2026anyview}), camera-controlled generation (ReCamMaster~\cite{bai2025recammaster}, CameraAnything~\cite{li2026cameraanything}), and exo-to-ego translation (WorldWander~\cite{song2026worldwander}, EgoX~\cite{kang2026egox}).
As TrajectoryCrafter provides no training code, we evaluated its released checkpoint.
All other baselines were fine-tuned on CoMind from two initialisations, either the generic video backbone or the authors' released cross-view pretrained weights, to evaluate cross-view pretraining.\footnote{The released CameraAnything checkpoint is pretrained with pinhole cameras; fine-tuning it on the fisheye videos of CoMind produced non-finite outputs, so we report only the version initialised from Wan2.1.}

\noindent\textbf{Evaluation Metrics.}
Following prior work on cross-view and egocentric video generation~\cite{van2026anyview,kang2026egox,song2026worldwander,bai2025recammaster}, we evaluated frame-level visual quality with PSNR, SSIM~\cite{wang2004ssim}, and LPIPS~\cite{zhang2018lpips}, semantic similarity with frame-wise CLIP~\cite{radford2021clip} (CLIP-I) and video-level ViCLIP~\cite{wang2024internvid}, and video distribution quality with FVD~\cite{unterthiner2018fvd}.
To evaluate Level-2 perspective taking, we measured scene layout accuracy using the mean IoU across 16 structural scene categories segmented by SAM~3~\cite{carion2025sam3} (Layout mIoU, in percent).
To evaluate the collaborator's head and hand motion, we followed AnyView~\cite{van2026anyview} to measure translation (Transl, in metres) and rotation (Rot, in degrees) errors between camera trajectories estimated by MapAnything~\cite{keetha2025mapanything} on the generated and ground-truth videos, and followed E3C~\cite{gu2026e3c} to measure the mean IoU of SAM~3 collaborator hand masks (Hand-mIoU, in percent).

\noindent\textbf{Implementation Details.}
We built \methodName on three backbones: two general-purpose video models, Wan2.2-5B and Wan2.1-14B~\cite{wan2025}, and AnyView~\cite{van2026anyview}, which pretrains Cosmos-Predict2-2B~\cite{nvidia2025cosmospredict2} for dynamic view synthesis on multi-view data.
For a fair comparison, we followed the training setups of WorldWander~\cite{song2026worldwander} on Wan2.2-5B, EgoX~\cite{kang2026egox} on Wan2.1-14B, and AnyView~\cite{van2026anyview} on its pretrained backbone, training all models for 5 epochs (details in the supplementary material).

\subsection{Comparison to Baselines}
\label{sec:comparison}

\begin{figure*}[t]
\centering
\includegraphics[width=\textwidth]{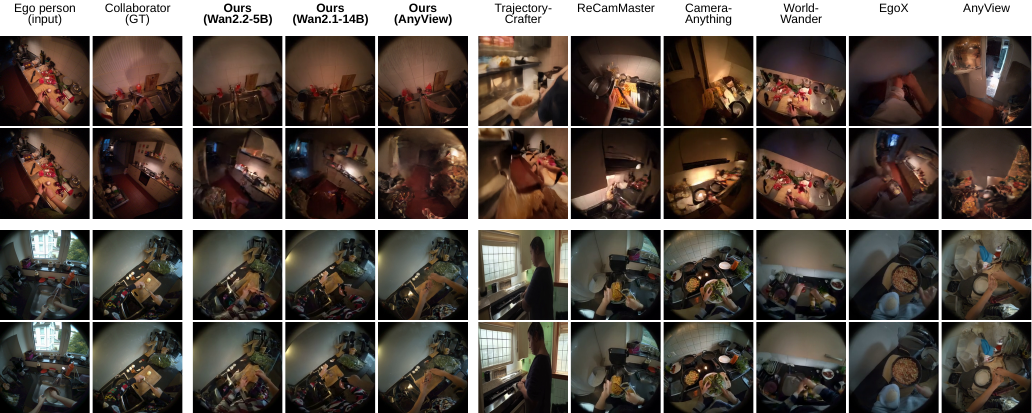}
\caption{\textbf{Qualitative comparison.} Two test examples, each at two time steps. For each baseline, we show the version with the higher PSNR in \cref{tab:main}.}
\label{fig:qual}
\end{figure*}

\cref{tab:main} reports quantitative results on CoMind.
Across all three backbones, \methodName outperformed the baselines built on the same backbone on all metrics.
Compared to the strongest baseline on each backbone in each metric, \methodName improved PSNR by $2.50$\,dB over WorldWander on Wan2.2-5B, $1.19$\,dB over EgoX on Wan2.1-14B, and $1.34$\,dB over AnyView, while reducing LPIPS by 23\% to 30\% and FVD by 22\% to 26\%.
\methodName substantially improved Level-2 scene layout, increasing Layout mIoU from at most $15.9$ among the baselines to $20.7$ on Wan2.2-5B, $26.1$ on Wan2.1-14B, and $23.6$ on AnyView, suggesting that shared 3D scene geometry helps reconstruct the collaborator's scene layout.
It also better captured the collaborator's head and hand motion, reducing Transl and Rot and improving Hand-mIoU from $17.5$ to $24.9$ on Wan2.2-5B, $21.2$ to $26.9$ on Wan2.1-14B, and $27.1$ to $30.9$ on AnyView.

Comparing the two initialisations of each baseline in \cref{tab:main} shows the effect of cross-view pretraining.
On AnyView, multi-view pretraining with Plücker rays improved PSNR from $16.32$ to $18.23$\,dB and reduced FVD from $1196.4$ to $228.0$.
However, even without any cross-view pretraining, \methodName on the general-purpose Wan2.2-5B and Wan2.1-14B backbones outperformed the cross-view pretrained AnyView on most metrics, obtaining higher PSNR ($18.57$ and $18.89$ vs.\ $18.23$\,dB), higher SSIM ($0.538$ and $0.551$ vs.\ $0.513$), lower LPIPS ($0.356$ and $0.343$ vs.\ $0.400$), lower FVD ($181.2$ and $164.7$ vs.\ $228.0$), lower translation error ($0.32$ and $0.28$ vs.\ $0.43$\,m), and higher Layout mIoU ($20.7$ and $26.1$ vs.\ $15.9$).
This suggests that explicit 3D scene geometry and ego-person visual memory provide stronger cues for visual perspective taking than multi-view pretraining alone.
Moreover, adding \methodName to pretrained AnyView further improved PSNR by $1.34$\,dB (to $19.57$\,dB), reduced LPIPS by 24\% (to $0.306$), and increased Layout mIoU to $23.6$ and Hand-mIoU to $30.9$, showing that our modules and cross-view pretraining are complementary.

\cref{fig:qual} shows qualitative comparisons on two examples where the two collaborators look at different parts of the kitchen.
TrajectoryCrafter warps the ego person's video using depth estimated only from that input view; since it cannot render what the ego person does not observe, it failed to follow the collaborator's camera ($23.5^{\circ}$ rotation error) and hallucinated content (\eg, a person by the window in the bottom example in \cref{fig:qual}).
Camera-controlled baselines condition on the target trajectory without shared 3D scene geometry, while exo-to-ego baselines assume that the input camera sees the target person, which often does not hold between collaborators.
Without a shared spatial reference and past ego-person frames, these baselines either copied content from the ego person's view (\eg, WorldWander repeating the ego person's counter in the top example of \cref{fig:qual}) or generated kitchen layouts that did not match the collaborator's viewpoint (Layout mIoU $\le 15.7$).
In contrast, \methodName placed scene structures such as the sink and wall-mounted cutting board at the correct locations across all three backbones.
Additional qualitative and video results are in the supplementary material.

\begin{table}[t]
\centering
\small
\setlength{\tabcolsep}{3.5pt}
\resizebox{\columnwidth}{!}{%
\begin{tabular}{@{}llcccccc@{}}
\toprule
 & & \multicolumn{2}{c}{\textbf{Never (46\%)}} & \multicolumn{2}{c}{\textbf{Partly (34\%)}} & \multicolumn{2}{c}{\textbf{Mostly (20\%)}} \\
\textbf{Method} & \textbf{Backbone} & \textbf{PSNR}$\uparrow$ & \textbf{SSIM}$\uparrow$ & \textbf{PSNR}$\uparrow$ & \textbf{SSIM}$\uparrow$ & \textbf{PSNR}$\uparrow$ & \textbf{SSIM}$\uparrow$ \\
\midrule
WorldWander~\cite{song2026worldwander} & Wan2.2-5B & 15.61 & 0.435 & 16.34 & 0.459 & 16.63 & 0.471 \\
EgoX~\cite{kang2026egox} & Wan2.1-14B & 17.45 & 0.516 & 17.86 & 0.529 & 18.01 & 0.540 \\
AnyView~\cite{van2026anyview} & AnyView & 17.68 & 0.494 & 18.66 & 0.525 & 18.74 & 0.532 \\
\midrule
\multirow{3}{*}{\textbf{\methodName (ours)}} & Wan2.2-5B & 18.35 & 0.531 & 18.78 & 0.543 & 18.68 & 0.545 \\
 & Wan2.1-14B & \underline{18.75} & \underline{0.548} & \underline{19.09} & \underline{0.554} & \underline{18.90} & \underline{0.551} \\
 & AnyView & \textbf{19.30} & \textbf{0.549} & \textbf{19.85} & \textbf{0.562} & \textbf{19.71} & \textbf{0.561} \\
\bottomrule
\end{tabular}}
\caption{\textbf{Results by collaborator visibility in the ego person's view} (never: $0$, partly: $1\text{--}24$, mostly: $\ge 25$ of $49$ frames; see the supplementary material for all baselines and LPIPS).}
\label{tab:vis}
\end{table}

\cref{tab:vis} groups the test clips by how often the collaborator's head appears in the ego person's view.
When the collaborator's head never appeared in the ego person's view ($46\%$ of clips), the strongest baselines on each backbone dropped by $0.56$ to $1.06$\,dB in PSNR relative to the mostly visible subset ($15.61$ vs.\ $16.63$\,dB for WorldWander, $17.45$ vs.\ $18.01$\,dB for EgoX, and $17.68$ vs.\ $18.74$\,dB for AnyView).
In contrast, \methodName dropped by only $0.15$ to $0.41$\,dB, achieving $18.35$, $18.75$, and $19.30$\,dB PSNR on the never-visible clips and improving over the same-backbone baselines by $2.74$, $1.30$, and $1.62$\,dB.

\subsection{Ablation Study}
\label{sec:ablation}

\begin{table*}[t]
\centering
\small
\setlength{\tabcolsep}{3.2pt}
\resizebox{\textwidth}{!}{%
\begin{tabular}{@{}llcccccccccc@{}}
\toprule
\textbf{Backbone} & \textbf{Variant} & \textbf{PSNR}$\uparrow$ & \textbf{SSIM}$\uparrow$ & \textbf{LPIPS}$\downarrow$ & \textbf{CLIP-I}$\uparrow$ & \textbf{ViCLIP}$\uparrow$ & \textbf{FVD}$\downarrow$ & \textbf{Transl}$\downarrow$ & \textbf{Rot}$\downarrow$ & \textbf{Layout}$\uparrow$ & \textbf{Hand}$\uparrow$ \\
\midrule
\multirow{6}{*}{Wan2.2-5B} & Base model & 16.19 & 0.460 & 0.506 & 0.811 & 0.819 & 281.0 & 0.40 & 16.1 & 7.2 & 20.7 \\
 & + DGR & 17.58 & 0.511 & 0.402 & 0.833 & 0.855 & 236.3 & 0.34 & 10.5 & 15.1 & 24.5 \\
 & + RGA & 17.89 & 0.517 & 0.386 & 0.837 & 0.865 & 208.8 & \textbf{0.32} & 9.4 & 18.3 & 24.6 \\
 & + Memory (full model) & \textbf{18.57} & \textbf{0.538} & \textbf{0.356} & \textbf{0.844} & 0.879 & \textbf{181.2} & \textbf{0.32} & \textbf{9.0} & \textbf{20.7} & \textbf{24.9} \\
\cmidrule{2-12}
 & Full model, sparse points & 18.42 & 0.533 & 0.369 & 0.842 & 0.879 & 192.8 & 0.34 & 9.1 & 20.4 & 23.5 \\
 & Full model, depth maps & 18.38 & 0.533 & 0.363 & \textbf{0.844} & \textbf{0.880} & 184.4 & 0.33 & \textbf{9.0} & 20.5 & \textbf{24.9} \\
\midrule
\multirow{6}{*}{Wan2.1-14B} & Base model & 16.59 & 0.498 & 0.525 & 0.789 & 0.710 & 727.0 & 0.34 & 19.2 & 3.8 & 20.4 \\
 & + DGR & 17.64 & 0.504 & 0.397 & 0.834 & 0.862 & 244.5 & 0.33 & 11.4 & 19.5 & 26.4 \\
 & + RGA & 18.38 & 0.529 & 0.365 & 0.850 & \textbf{0.899} & 191.6 & 0.30 & 10.0 & 25.1 & 26.0 \\
 & + Memory (full model) & \textbf{18.89} & \textbf{0.551} & \textbf{0.343} & \textbf{0.853} & 0.897 & \textbf{164.7} & \textbf{0.28} & 10.1 & \textbf{26.1} & \textbf{26.9} \\
\cmidrule{2-12}
 & Full model, sparse points & 18.19 & 0.529 & 0.369 & 0.851 & 0.897 & 205.1 & 0.29 & \textbf{9.9} & 23.2 & 24.7 \\
 & Full model, depth maps & 17.66 & 0.516 & 0.367 & 0.844 & 0.879 & 222.2 & \textbf{0.28} & \textbf{9.9} & 23.2 & \textbf{26.9} \\
\midrule
\multirow{6}{*}{AnyView} & Base model & 18.23 & 0.513 & 0.400 & 0.872 & 0.886 & 228.0 & 0.43 & 9.3 & 15.9 & 27.1 \\
 & + DGR & 18.91 & 0.536 & 0.347 & 0.876 & 0.894 & 219.1 & 0.44 & 9.1 & 19.1 & 29.6 \\
 & + RGA & 18.97 & 0.538 & 0.346 & 0.876 & 0.893 & 214.7 & 0.44 & 9.1 & 19.1 & 30.1 \\
 & + Memory (full model) & \textbf{19.57} & \textbf{0.556} & \textbf{0.306} & \textbf{0.888} & \textbf{0.916} & \textbf{177.1} & \textbf{0.41} & 8.8 & \textbf{23.6} & \textbf{30.9} \\
\cmidrule{2-12}
 & Full model, sparse points & 19.03 & 0.539 & 0.337 & 0.885 & 0.908 & 193.4 & 0.43 & 8.8 & 21.2 & 28.9 \\
 & Full model, depth maps & 19.19 & 0.544 & 0.325 & 0.886 & 0.910 & 190.0 & 0.42 & \textbf{8.7} & 21.6 & 29.9 \\
\bottomrule
\end{tabular}}
\caption{\textbf{Ablations on the proposed modules.} Starting from the base model, we add DGR, RGA, and the ego person's visual memory one at a time. The last two rows replace the scene coordinate maps of the full model with sparse points or depth maps. Bold: best per backbone.}
\label{tab:ablation}
\end{table*}

\cref{tab:ablation} evaluates the contribution of each module across all three backbones, starting from the Plücker-conditioned base model (\cref{sec:backbone}) and adding dense geometric rendering (DGR), relative geometric attention (RGA), and the ego person's visual memory one by one.
Adding DGR brought the largest single PSNR improvement on all three backbones, increasing PSNR by $1.39$\,dB on Wan2.2-5B, $1.05$\,dB on Wan2.1-14B, and $0.68$\,dB on AnyView.
By representing all views in a shared 3D coordinate system, DGR reduced camera rotation error by 35\% and 41\% on the two Wan backbones and more than doubled their Layout mIoU (from $7.2$ to $15.1$ and from $3.8$ to $19.5$).
To test whether this improvement came from dense, view-invariant world coordinates rather than 3D geometry alone, we replaced the scene coordinate maps in the full model with either sparse uncoloured point projections (\emph{Full model, sparse points}) or camera-space depth maps (\emph{Full model, depth maps}) rendered from the same point cloud.
Sparse point projection lowered PSNR by $0.15$ to $0.70$\,dB, worsened FVD (\eg, $164.7$ to $205.1$ on Wan2.1-14B), and reduced Layout mIoU and Hand-mIoU, while depth maps reduced PSNR by $0.19$ to $1.23$\,dB and lowered Layout mIoU (\eg, $26.1$ to $23.2$ on Wan2.1-14B and $23.6$ to $21.6$ on AnyView), confirming the importance of dense, view-invariant 3D coordinates.

Adding RGA on top of DGR further improved performance on the two general-purpose backbones, increasing PSNR by $0.31$ and $0.74$\,dB, reducing FVD by 12\% and 22\%, and improving Layout mIoU from $15.1$ to $18.3$ on Wan2.2-5B and from $19.5$ to $25.1$ on Wan2.1-14B.
On AnyView, RGA also reduced FVD from $219.1$ to $214.7$ and improved Hand-mIoU from $29.6$ to $30.1$, with a smaller PSNR gain ($+0.06$\,dB).
Because AnyView was pretrained with Plücker ray maps on 12 multi-view datasets~\cite{van2026anyview}, its backbone may already capture basic cross-view ray relationships from data, whereas general-purpose backbones benefited much more from the explicit $SE(3)$ ray-frame transform in RGA.

Finally, adding the ego person's visual memory gave the full model, increasing PSNR by $0.68$, $0.51$, and $0.60$\,dB, reducing FVD by 13\% to 18\%, and improving Layout mIoU to $20.7$, $26.1$, and $23.6$ on Wan2.2-5B, Wan2.1-14B, and AnyView, showing that multi-view pretraining cannot replace past observations outside the current view.
Moreover, on the 29 session-start test clips without earlier ego-person frames, our full model (trained with memory dropout) still outperformed the strongest baseline on every backbone ($18.10$ vs.\ $16.22$\,dB on Wan2.2-5B, $18.09$ vs.\ $17.75$\,dB on Wan2.1-14B, and $18.70$ vs.\ $18.03$\,dB on AnyView).

\subsection{Limitations and Future Work}
\label{sec:limitations}

First, because the ego person's egocentric camera never observes the ego person themselves, \methodName has no visual reference for the ego person's appearance when they enter the collaborator's view.
Second, surfaces absent from the scene point cloud, such as window glass, appear as empty regions in the scene coordinate maps, which \methodName may fill with incorrect content (see the supplementary material for both failure cases).
Third, because the scene point cloud is static, environment changes made during collaboration are not captured in the scene coordinate maps.
Fourth, as we trained and evaluated only on indoor scenes, the generalisation of \methodName to outdoor environments remains untested.
Promising future directions include combining our modules with cross-view pretraining on larger backbones and applying \methodName to online robot planning and multi-perspective policy learning, such as synthesising what a collaborator would see along candidate counterfactual trajectories (see the supplementary material for qualitative results).

\section{Conclusion}
\label{sec:conclusion}

We presented \methodName -- a method for \textit{generative visual perspective taking} that synthesises a collaborator's egocentric video from an ego person's video, a static scene point cloud, and the collaborator's camera trajectory.
By combining dense geometric rendering, relative geometric attention, and the ego person's visual memory, \methodName connects the two views through the 3D scene geometry shared by both collaborators.
Experiments across three backbones show that \methodName consistently improves visual quality, Level-2 scene layout accuracy, and collaborator motion over same-backbone baselines and complements cross-view pretraining.

\section*{Acknowledgements}
The authors gratefully acknowledge the scientific support and HPC resources provided by the Erlangen National High Performance Computing Center (NHR@FAU) of the Friedrich-Alexander-Universität Erlangen-Nürnberg (FAU) under the NHR project Project a113cb | JA-29757. NHR funding is provided by federal and Bavarian state authorities.
{
    \small
    \bibliographystyle{ieeenat_fullname}
    \bibliography{main}
}

\clearpage
\setcounter{page}{1}
\maketitlesupplementary
\appendix
\section{Method Details}
\label{sec:supp_method}

\subsection{Base Model}
\label{sec:supp_base}

\paragraph{Plücker ray maps.}
Plücker ray maps have proven effective for scene representation~\cite{sitzmann2021lfn}, camera-controlled generation~\cite{he2025cameractrl,li2026cameraanything}, dynamic view synthesis~\cite{van2026anyview}, and policy learning~\cite{jiang2026knowyourcamera}.
For pixel $\mathbf{u}$ of frame $l$ with camera-to-world pose $T_{l}=[R_{l}\,|\,\mathbf{c}_{l}]$, where $\mathbf{c}_{l}$ is the camera centre, and intrinsics $K$, the ray direction is $\mathbf{d}_{l}(\mathbf{u})=R_{l}\,\pi^{-1}(\mathbf{u};K)/\lVert\pi^{-1}(\mathbf{u};K)\rVert$, where $\pi^{-1}$ unprojects a pixel to its viewing ray, and the Plücker embedding is
\begin{equation}
\mathbf{r}_{l}(\mathbf{u})=\big(\mathbf{d}_{l}(\mathbf{u}),\ \mathbf{c}_{l}\times\mathbf{d}_{l}(\mathbf{u})\big)\in\mathbb{R}^{6}.
\label{eq:plucker}
\end{equation}

\paragraph{Motion term.}
LaMo~\cite{jiang2026lamo} shows that supervising the global latent motion of the predicted clean video leads to more physically realistic motion in video generation.
Following LaMo, we compute the clean estimate $\hat{\mathbf{z}}^{c}_{0}=\mathbf{z}^{c}_{t}-t\,\mathbf{v}_{\theta}$.
Let $\boldsymbol{\mu}_{f}(\mathbf{z})$ be the spatial mean of the latent difference between frames $f+1$ and $f$; the motion term is
\begin{equation}
\mathcal{L}_{\text{mot}}=\big(1-t^{2}\big)\,\frac{1}{F-1}\sum_{f}\frac{\lVert\boldsymbol{\mu}_{f}(\hat{\mathbf{z}}^{c}_{0})-\boldsymbol{\mu}_{f}(\mathbf{z}^{c}_{0})\rVert_{2}^{2}}{\lVert\boldsymbol{\mu}_{f}(\mathbf{z}^{c}_{0})\rVert_{2}^{2}},
\label{eq:mot}
\end{equation}
where $F$ is the number of latent frames, and the weight $1-t^{2}$ reduces the influence of high noise levels, at which $\hat{\mathbf{z}}^{c}_{0}$ is unreliable.

\paragraph{Conditioning inputs.}
Each six-dimensional Plücker map is split into its direction and moment components, which are stored as two three-channel videos and encoded separately by the frozen VAE; camera centres are expressed in the frame of the collaborator's first camera and divided by the scene scale $s$.
The clean ego-person tokens, including the memory tokens, are assigned diffusion time $t=0$.
We keep the positional encoding of each backbone.

\subsection{Ray-Aligned Frames}
\label{sec:supp_rga}

We follow the ray-aligned frame construction of UCPE~\cite{zhang2026unified} and extend the attention head to cross-person perspective taking.
For token $i$ of a frame with camera-to-world pose $T=[R\,|\,\mathbf{c}]$, let $\mathbf{e}^{z}_{i}$ be the unit world-space viewing ray through the token centre, obtained by unprojecting the token centre with the camera model of the device.
The $x$-axis is $\mathbf{e}^{x}_{i}=\mathbf{e}_{\text{up}}\times\mathbf{e}^{z}_{i}/\lVert\mathbf{e}_{\text{up}}\times\mathbf{e}^{z}_{i}\rVert$, where $\mathbf{e}_{\text{up}}$ is the world-space vertical axis of the camera, \ie, the second column of $R$; if the ray is nearly parallel to $\mathbf{e}_{\text{up}}$, we use the horizontal camera axis instead.
With $\mathbf{e}^{y}_{i}=\mathbf{e}^{z}_{i}\times\mathbf{e}^{x}_{i}$ and $Q_{i}=[\mathbf{e}^{x}_{i}\,\mathbf{e}^{y}_{i}\,\mathbf{e}^{z}_{i}]$, the world-to-frame transform is
\begin{equation}
\Gamma_{i}=\begin{bmatrix}Q_{i}^{\top} & -Q_{i}^{\top}\mathbf{c}/s\\ \mathbf{0}^{\top} & 1\end{bmatrix},
\end{equation}
where the camera centre is divided by the scene scale $s$ of the scene coordinate maps in the main paper.
The branch has its own query, key, value, and output projections.
Unlike UCPE, which combines the ray transform with image-space RoPE over a single camera's video, each head in RGA splits its channels into two halves: a rotary encoding of the frame index shared by synchronised ego-person and collaborator frames, and geometric channels split into groups of four, on each of which $\Gamma_{i}\Gamma_{j}^{-1}$ acts in the relative geometric attention of the main paper.
In practice, we transform the query by $\Gamma_{i}^{\top}$ and the key and value by $\Gamma_{j}^{-1}$, so that standard attention kernels can be used, and transform the output back by $\Gamma_{i}$.
The branch is added to every fourth transformer block.

\subsection{Dense Geometric Rendering}
\label{sec:supp_dgr}

The scene coordinate maps are rendered at the video resolution of $256\times256$ with the camera model of the device.
To obtain dense maps from the point cloud, which has a voxel size of $v=2$\,cm, each point is splatted as a disk whose radius approximates its projected footprint, $r=\mathrm{clip}\big(\mathrm{round}(f\,v\,\sqrt{|\mathbf{n}^{\top}\mathbf{d}|}/\rho),1,4\big)$ pixels, where $f$ is the focal length in pixels, $\rho$ the distance of the point to the camera, $\mathbf{n}$ its normal, and $\mathbf{d}$ the viewing direction.
Depth testing keeps the point with the smallest $\rho$ per pixel, and points closer than 8\,cm to the image plane are discarded.
Coordinates are expressed in metres and normalised by the scene scale $s=8$\,m.
In addition to the VAE-encoded scene coordinate map, each view receives an occupancy map, \ie, the fraction of pixels in each latent cell that are covered by rendered points, which is concatenated at the latent resolution without passing through the VAE.

\subsection{Memory Frame Selection}
\label{sec:supp_memory}

Candidate memory frames are ego-person frames sampled at 2\,Hz before the clip.
The visible point sets $\mathcal{P}^{c}_{l}$ and $\mathcal{P}_{l'}$ in the coverage objective of the main paper are obtained by depth-tested point splatting into each view, and each visible point is counted once per collaborator frame.
Each collaborator frame in $\mathcal{L}$ contributes equally, independently of how many points it observes, and frames without visible points are skipped.
At each greedy step, we add the candidate with the largest gain in $f$; ties are broken by the pose distance to the collaborator, and then by recency.
After selection, the memory frames are sorted by time.
The memory tokens receive the temporal positions $-n,\dots,-1$ before the clip, where $n\le M$ is the number of selected frames, and each memory frame is conditioned on its own Plücker map and scene coordinate map.
If fewer than $M$ earlier ego-person frames are available, the missing memory slots are masked out of attention on Wan2.2-5B and AnyView and removed from the token sequence on Wan2.1-14B.
During training, we drop the whole memory with probability $0.15$, so that the model also learns to generate the collaborator's view without memory.

\section{Experimental Details}
\label{sec:supp_exp}

\subsection{Dataset}
\label{sec:supp_data}
CoMind~\cite{gavryushin2026comind} records pairs of participants wearing egocentric glasses that capture synchronised fisheye videos and camera poses alongside a scene point cloud per environment.
We extracted 49-frame clips at 10\,fps and resized the fisheye videos to $256\times256$.
Each clip was used in both directions, treating either participant as the ego person, which yielded 7{,}670 training, 24 validation, and 1{,}020 test clips across 15 held-out test sessions that are disjoint from the training sessions.
To quantify how often the collaborator appears in the ego person's view, we projected the collaborator's camera centre (head position) into the ego person's fisheye image at every frame: across the 1{,}020 test clips, the collaborator's head was never in view (0 frames) in 466 clips (46\%), partly in view (1 to 24 frames) in 348 clips (34\%), and in view in at least half of the frames ($\ge 25$ frames) in 206 clips (20\%).

\subsection{Baselines}
\label{sec:supp_baselines}
We adapted six representative methods from novel view synthesis (TrajectoryCrafter~\cite{yu2025trajectorycrafter} and AnyView~\cite{van2026anyview}), camera-controlled video generation (ReCamMaster~\cite{bai2025recammaster} and CameraAnything~\cite{li2026cameraanything}), and exo-to-ego translation (WorldWander~\cite{song2026worldwander} and EgoX~\cite{kang2026egox}).
Each baseline received the ego person's video as its source video and, whenever its interface accepts a target camera, the collaborator's camera trajectory to generate the collaborator's video.
All baselines retained their original architecture, camera representation, and default inference settings.
Except for TrajectoryCrafter, all baselines generated 49 frames at $256\times256$.
All methods that take a text prompt, including ours, used a fixed prompt.
Apart from TrajectoryCrafter and CameraAnything, we fine-tuned each baseline on the CoMind training split for 5 epochs under two initialisations: once starting from its generic video backbone and once from the cross-view pretrained weights released by its authors.

\paragraph{TrajectoryCrafter.}
Because its training code is not publicly available, we ran the released inference pipeline unchanged: DepthCrafter~\cite{hu2025depthcrafter} estimated per-frame depth from the ego person's video, which was warped into the collaborator's viewpoint and inpainted by CogVideoX-Fun-V1.1-5B-InP~\cite{yang2025cogvideox} using the released TrajectoryCrafter weights at $384\times672$.
Its camera interface parameterises the target trajectory via two spherical angles and a radial distance relative to the source camera; we converted the collaborator's poses relative to the ego person's first camera into this parameterisation and resized the generated videos to $256\times256$ for evaluation.

\paragraph{ReCamMaster.}
We retained its Wan2.1-T2V-1.3B backbone, camera extrinsics encoder, and frame-dimension concatenation of source and target tokens.
The camera condition consisted of the collaborator's $3\times4$ extrinsic matrices relative to the ego person's first camera sampled every fourth frame.
We trained for 38{,}350 steps with batch size 1 and learning rate $10^{-4}$, once initialised from Wan2.1 (with a zero-initialised camera encoder and identity-initialised projector) and once from the authors' released checkpoint.

\paragraph{CameraAnything.}
We used Wan2.1-T2V-14B with the official camera module, which injects Plücker ray maps via adaptive layer normalisation.
Source and target cameras were specified by camera-to-world poses and the horizontal field of view of a pinhole approximation to the fisheye lens.
We trained with learning rate $5\times10^{-5}$ and batch size 1 per GPU across four GPUs for 9{,}590 steps.
Fine-tuning the released pretrained checkpoint, whose camera encoder was trained on pinhole intrinsics, produced non-finite outputs, possibly due to the wide-angle fisheye geometry of CoMind; we therefore report the version fine-tuned from Wan2.1.

\paragraph{WorldWander.}
The ego person's video served as the input video; WorldWander does not accept target camera poses.
We trained rank-80 LoRA adapters on Wan2.2-TI2V-5B with batch size 4 and learning rate $5\times10^{-5}$, once initialised from Wan2.2 and once from the released exo-to-ego LoRA weights.

\paragraph{EgoX.}
The ego person's video and camera trajectory served as the source view, and the collaborator's trajectory served as the target egocentric camera.
Following the official EgoX pipeline, ViPE estimated depth from the ego person's video to lift a per-frame point cloud that was rendered into the collaborator's viewpoint as a geometric prior, combined with geometry-guided attention and width-wise token concatenation.
We trained rank-256 LoRA adapters on Wan2.1-I2V-14B with learning rate $2\times10^{-5}$, initialised either from Wan2.1 (batch size 1) or from the released EgoX checkpoint (batch size 2).

\paragraph{AnyView.}
The ego person's video served as the source view and the collaborator's video as the target view, with both views conditioned on pixel-wise Plücker ray maps anchored at the collaborator's first camera.
We fine-tuned all parameters with batch size 1 and learning rate $2\times10^{-6}$, once initialised from Cosmos-Predict2-2B and once from the dynamic view synthesis pretrained checkpoint (AnyView).

\subsection{Evaluation Metrics}
\label{sec:supp_metrics}
All metrics were evaluated over the 1{,}020 test clips against the ground-truth collaborator videos.
PSNR, SSIM~\cite{wang2004ssim}, LPIPS~\cite{zhang2018lpips}, and CLIP-I were averaged across all frames and clips, where CLIP-I is the cosine similarity between frame-level CLIP ViT-B/32~\cite{radford2021clip} embeddings.
ViCLIP~\cite{wang2024internvid} measured the cosine similarity between ViCLIP-L/14 video embeddings extracted from 8 uniformly sampled frames, and FVD~\cite{unterthiner2018fvd} was computed using I3D features.
For Transl and Rot, following the camera-pose evaluation of AnyView~\cite{van2026anyview}, MapAnything~\cite{keetha2025mapanything} estimated camera trajectories from the generated and ground-truth videos (each expressed relative to its initial frame), and we reported the mean Euclidean translation error (in metres) and geodesic rotation error (in degrees).
For Layout mIoU, SAM~3~\cite{carion2025sam3} segmented 16 structural scene categories (window, wall, ceiling, floor, door, cabinet, countertop, tiled backsplash, sink, stove, range hood, refrigerator, oven, microwave, shelf, and table) per clip, excluding pixels belonging to humans and hands; we computed the dataset-wide IoU per category and averaged across the 16 categories.
For Hand-mIoU, SAM~3 segmented hands and gloves per clip; we retained the collaborator's own hands (verified via ground-truth hand keypoints and excluding masks belonging to the ego person) and reported the frame-averaged IoU, where frames in which neither the ground truth nor the prediction contained hands were assigned an IoU of 1.

\subsection{Implementation Details}
\label{sec:supp_impl}
On Wan2.2-5B~\cite{wan2025}, we followed WorldWander~\cite{song2026worldwander} and trained rank-80 LoRA adapters alongside our geometric modules with batch size 4 and learning rate $5\times10^{-5}$.
On Wan2.1-14B~\cite{wan2025}, we followed EgoX~\cite{kang2026egox} and trained rank-256 LoRA adapters on the attention projections alongside our geometric modules with batch size 4 and learning rate $2\times10^{-5}$.
On AnyView~\cite{van2026anyview} (Cosmos-Predict2-2B~\cite{nvidia2025cosmospredict2} pretrained for dynamic view synthesis), we followed AnyView and fine-tuned all parameters with batch size 1 and learning rate $2\times10^{-6}$.
All models were trained for 5 epochs, using a single NVIDIA H100 GPU for Wan2.2-5B and AnyView and four H100 GPUs for Wan2.1-14B.
Newly added input channels and the output projections of relative geometric attention were zero-initialised, and the video VAE remained frozen.
Relative geometric attention was inserted into every fourth transformer block.
Historical memory frames ($M=4$) were selected from ego-person frames sampled at 2\,Hz prior to the clip.
The metric scene scale was set to $s=8$\,m, and the motion consistency weight was $\lambda=0.4$.
During training, the geometric renderings of the ego person, the collaborator, and the memory frames were dropped jointly with probability $0.15$.
At inference, this enabled classifier-free guidance on the geometry: we combined the network outputs (velocity on the Wan backbones, clean sample on AnyView) with and without geometry as $\mathbf{y}_{\varnothing}+w\,(\mathbf{y}-\mathbf{y}_{\varnothing})$ with $w=1.5$, where $\mathbf{y}_{\varnothing}$ was predicted with the collaborator's geometric rendering set to zero on Wan2.2-5B and AnyView, and with all geometric renderings of the ego person, the collaborator, and the memory set to zero on Wan2.1-14B.
We used 50 denoising steps on the Wan backbones and 35 steps on AnyView.

\subsection{Additional Ablations and Analysis}
\label{sec:supp_plucker_motion}

\paragraph{Pl\"ucker ray maps and motion consistency.}
Because AnyView requires full-parameter fine-tuning and training LoRA on Wan2.1-14B is also slow, we did not have sufficient computational resources to run every design ablation on all three backbones. We therefore conducted the additional ablations below on Wan2.2-5B and transferred the resulting design choices directly to Wan2.1-14B and AnyView.

Our base model conditions each view on Pl\"ucker ray maps, and the full model incorporates the latent motion consistency loss $\mathcal{L}_{\text{mot}}$.
We evaluated these two design choices on the Wan2.2-5B backbone in \cref{tab:supp_ww}.
Adding Pl\"ucker ray conditioning to WorldWander reduced camera rotation error from $25.0^{\circ}$ to $16.1^{\circ}$ while improving PSNR, SSIM, LPIPS, and FVD.
In the full model, including the motion consistency term $\mathcal{L}_{\text{mot}}$ reduced FVD from $189.1$ to $181.2$, lowered translation error from $0.35$ to $0.32$\,m, and added $0.08$\,dB in PSNR.

\begin{table*}[t]
\centering
\small
\setlength{\tabcolsep}{3.2pt}
\resizebox{\textwidth}{!}{%
\begin{tabular}{@{}lcccccccccc@{}}
\toprule
\textbf{Variant} & \textbf{PSNR}$\uparrow$ & \textbf{SSIM}$\uparrow$ & \textbf{LPIPS}$\downarrow$ & \textbf{CLIP-I}$\uparrow$ & \textbf{ViCLIP}$\uparrow$ & \textbf{FVD}$\downarrow$ & \textbf{Transl}$\downarrow$ & \textbf{Rot}$\downarrow$ & \textbf{Layout}$\uparrow$ & \textbf{Hand}$\uparrow$ \\
\midrule
WorldWander (no camera condition) & 15.94 & 0.449 & 0.516 & \textbf{0.811} & \textbf{0.824} & 288.7 & 0.48 & 25.0 & 6.1 & 17.5 \\
+ Pl\"ucker ray maps & \textbf{16.19} & \textbf{0.460} & \textbf{0.506} & \textbf{0.811} & 0.819 & \textbf{281.0} & \textbf{0.40} & \textbf{16.1} & \textbf{7.2} & \textbf{20.7} \\
\midrule
Full model without the motion term & 18.49 & 0.536 & \textbf{0.356} & 0.842 & 0.878 & 189.1 & 0.35 & \textbf{9.0} & 20.6 & \textbf{25.1} \\
Full model & \textbf{18.57} & \textbf{0.538} & \textbf{0.356} & \textbf{0.844} & \textbf{0.879} & \textbf{181.2} & \textbf{0.32} & \textbf{9.0} & \textbf{20.7} & 24.9 \\
\bottomrule
\end{tabular}}
\caption{\textbf{Pl\"ucker ray maps and motion term on Wan2.2-5B.} Bold: better per block.}
\label{tab:supp_ww}
\end{table*}

\begin{figure*}[t]
\centering
\includegraphics[width=\textwidth]{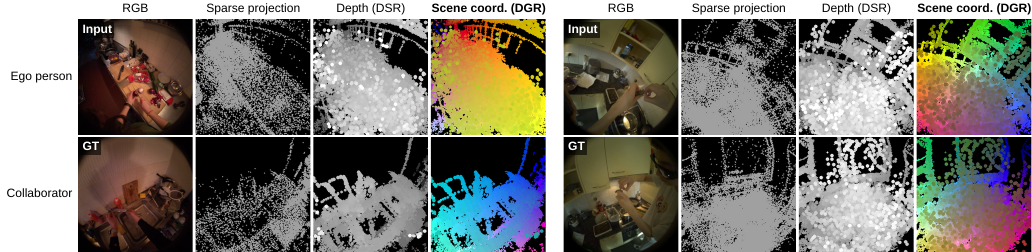}
\caption{\textbf{Geometric conditions.} Ego person and collaborator views of two test clips at the same time step. The collaborator's RGB frame (GT) is shown only for reference. Black pixels are not covered by the point cloud. Scene coordinate colours are contrast-stretched for display with one linear map per clip.}
\label{fig:supp_cond}
\end{figure*}

\paragraph{Visualising geometric representations.}
\label{sec:supp_splat}
\cref{fig:supp_cond} shows the three geometric representations compared in the ablation study of the main paper: sparse point projection, camera-space depth maps, and our dense scene coordinate maps (DGR).
The sparse projection only marks where points fall and leaves gaps between them.
The depth maps fill the area around each point, but the same surface point takes different values in the ego person's and the collaborator's views because depth depends on the camera.
The scene coordinate maps also fill the area around each point and encode each point by its position in the shared frame, so the point has the same colour in every view.

\paragraph{Performance stratified by collaborator visibility.}
\label{sec:supp_vis}
\cref{tab:supp_vis} reports test results across clips in which the collaborator's head was never in view ($46\%$), partly in view ($34\%$), or mostly in view ($\ge 50\%$ of frames, $20\%$) in the ego person's video.
We observed two patterns.
First, when the collaborator's head was never in the ego person's view, the baselines dropped by up to $1.37$\,dB in PSNR compared with the mostly visible subset, whereas \methodName dropped by only $0.15$ to $0.41$\,dB, less than every baseline on the same backbone ($0.56$ to $1.06$\,dB), and improved PSNR over the strongest same-backbone baselines by $2.74$\,dB on Wan2.2-5B, $1.30$\,dB on Wan2.1-14B, and $1.62$\,dB on AnyView.
Second, in the mostly visible subset, where the two participants often face each other and the ego person occupies a large part of the collaborator's view, LPIPS was slightly higher than in the never-visible subset (\eg, $0.363$ vs.\ $0.338$ on Wan2.1-14B).
This likely occurs because the ego person's egocentric camera never observes the ego person's own face and torso, leaving their exact appearance ambiguous when they are visible in the collaborator's view.

\begin{table*}[t]
\centering
\small
\setlength{\tabcolsep}{3.5pt}
\resizebox{\textwidth}{!}{%
\begin{tabular}{@{}lllcccccccccc@{}}
\toprule
 & & & & \multicolumn{3}{c|}{\textbf{Never visible ($n=466$)}} & \multicolumn{3}{c|}{\textbf{Partly visible ($n=348$)}} & \multicolumn{3}{c}{\textbf{Mostly visible ($n=206$)}} \\
\textbf{Method} & \textbf{Task} & \textbf{Backbone} & \textbf{CV} & \textbf{PSNR}$\uparrow$ & \textbf{SSIM}$\uparrow$ & \textbf{LPIPS}$\downarrow$ & \textbf{PSNR}$\uparrow$ & \textbf{SSIM}$\uparrow$ & \textbf{LPIPS}$\downarrow$ & \textbf{PSNR}$\uparrow$ & \textbf{SSIM}$\uparrow$ & \textbf{LPIPS}$\downarrow$ \\
\midrule
TrajectoryCrafter$^{\dagger}$~\cite{yu2025trajectorycrafter} & NVS & CogVideoX-5B & \cmark & 13.04 & 0.300 & 0.640 & 13.18 & 0.315 & 0.641 & 13.26 & 0.337 & 0.649 \\
\cmidrule{1-13}
\multirow{2}{*}{ReCamMaster~\cite{bai2025recammaster}} & \multirow{2}{*}{Camera} & \multirow{2}{*}{Wan2.1-1.3B} & \xmark & 15.14 & 0.418 & 0.523 & 15.92 & 0.445 & 0.502 & 16.28 & 0.460 & 0.503 \\
 & & & \cmark & 16.12 & 0.469 & 0.493 & 16.94 & 0.496 & 0.476 & 17.49 & 0.516 & 0.470 \\
\cmidrule{1-13}
CameraAnything$^{\ddagger}$~\cite{li2026cameraanything} & Camera & Wan2.1-14B & \xmark & 16.31 & 0.466 & 0.472 & 16.91 & 0.485 & 0.451 & 17.22 & 0.499 & 0.453 \\
\cmidrule{1-13}
\multirow{2}{*}{WorldWander~\cite{song2026worldwander}} & \multirow{2}{*}{Exo-to-ego} & \multirow{2}{*}{Wan2.2-5B} & \xmark & 15.47 & 0.433 & 0.524 & 16.22 & 0.457 & 0.508 & 16.53 & 0.469 & 0.510 \\
 & & & \cmark & 15.61 & 0.435 & 0.515 & 16.34 & 0.459 & 0.495 & 16.63 & 0.471 & 0.499 \\
\cmidrule{1-13}
\multirow{2}{*}{EgoX~\cite{kang2026egox}} & \multirow{2}{*}{Exo-to-ego} & \multirow{2}{*}{Wan2.1-14B} & \xmark & 17.45 & 0.516 & 0.447 & 17.86 & 0.529 & 0.443 & 18.01 & 0.540 & 0.451 \\
 & & & \cmark & 17.28 & 0.501 & 0.464 & 17.74 & 0.519 & 0.446 & 17.89 & 0.532 & 0.452 \\
\cmidrule{1-13}
\multirow{2}{*}{AnyView~\cite{van2026anyview}} & \multirow{2}{*}{NVS} & \multirow{2}{*}{Cosmos-2B} & \xmark & 16.04 & 0.454 & 0.566 & 16.47 & 0.468 & 0.557 & 16.71 & 0.484 & 0.555 \\
 & & & \cmark & 17.68 & 0.494 & 0.427 & 18.66 & 0.525 & 0.374 & 18.74 & 0.532 & 0.382 \\
\midrule
\multirow{3}{*}{\textbf{\methodName (ours)}} & \multirow{3}{*}{Ego-to-ego} & Wan2.2-5B & \xmark & 18.35 & 0.531 & 0.359 & 18.78 & 0.543 & 0.346 & 18.68 & 0.545 & 0.367 \\
 & & Wan2.1-14B & \xmark & \underline{18.75} & \underline{0.548} & \underline{0.338} & \underline{19.09} & \underline{0.554} & \underline{0.337} & \underline{18.90} & \underline{0.551} & \underline{0.363} \\
 & & AnyView & \cmark & \textbf{19.30} & \textbf{0.549} & \textbf{0.310} & \textbf{19.85} & \textbf{0.562} & \textbf{0.295} & \textbf{19.71} & \textbf{0.561} & \textbf{0.313} \\
\bottomrule
\end{tabular}}
\caption{\textbf{Performance stratified by collaborator visibility in the ego person's video.} Test clips are grouped by the number of frames in which the collaborator's head centre projects into the ego person's view: never visible ($0$ frames, $46\%$), partly visible ($1\text{--}24$ frames, $34\%$), and mostly visible ($\ge 25$ of $49$ frames, $20\%$). Bold: best; underlined: second best.}
\label{tab:supp_vis}
\end{table*}

\subsection{Additional Qualitative Results}
\label{sec:supp_qual}
\begin{figure*}[p]
\centering
\includegraphics[width=\textwidth]{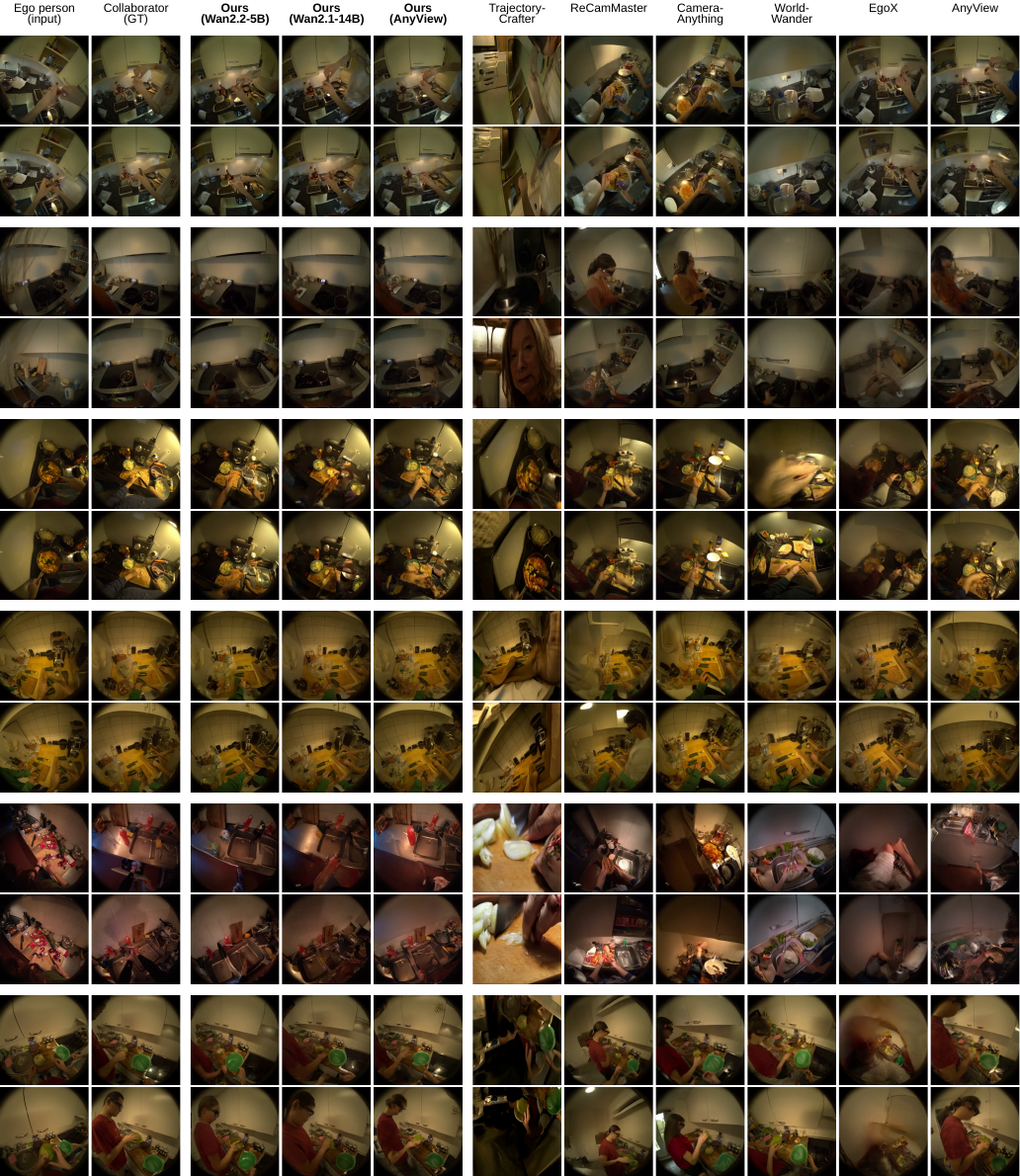}
\caption{\textbf{Additional qualitative comparison.} Six test clips, each at two time steps (top and bottom rows). For each baseline, we show the version with the higher PSNR in Table~1 of the main paper.}
\label{fig:supp_qual}
\end{figure*}

\cref{fig:supp_qual} compares \methodName with the baselines on six additional test clips from different sessions, following the layout of Figure~3 in the main paper.
Across these examples, baselines without shared 3D geometry and ego-person memory frequently copied content from the ego person's view or hallucinated mismatched room layouts.
For instance, in the second example, no person is visible in the collaborator's ground-truth view, yet ReCamMaster, CameraAnything, and AnyView hallucinated a person in the first frame and TrajectoryCrafter generated a face in the second frame.
In the fifth example, the ego person works at the counter while the collaborator faces the sink: in the first frame, all three \methodName models reconstructed the double sink with the red bottles behind it as in the ground truth, whereas TrajectoryCrafter stayed on a cutting board, CameraAnything and EgoX omitted the sink, and ReCamMaster and WorldWander rendered the sink from an inaccurate viewpoint.

\subsection{Counterfactual Trajectories}
\label{sec:supp_cf}
\begin{figure*}[t]
\centering
\includegraphics[width=\textwidth]{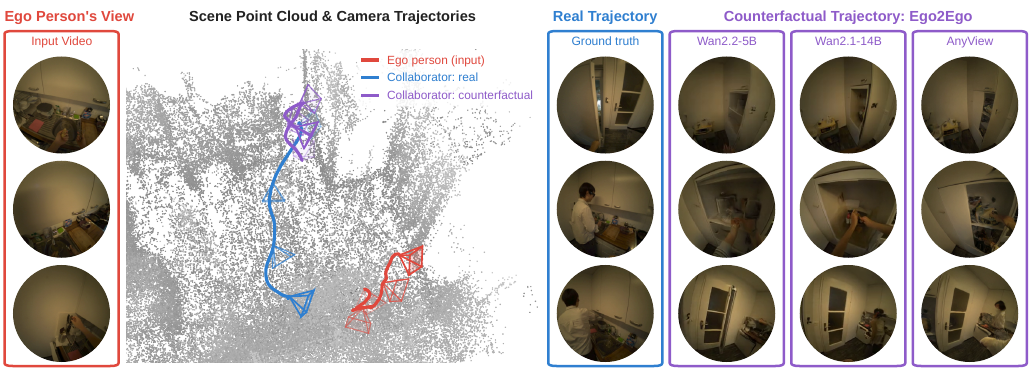}\\[4pt]
\includegraphics[width=\textwidth]{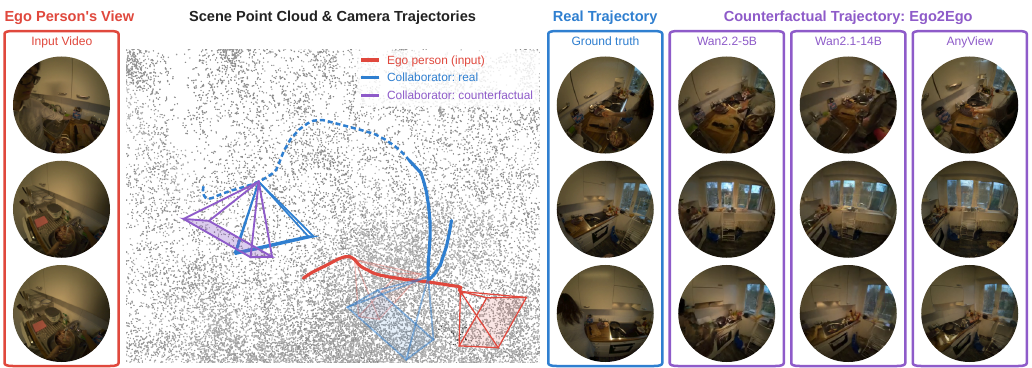}
\caption{\textbf{Counterfactual trajectories.} Top: the collaborator stays outside the ego person's view; rows show frames from the start, middle, and end of the clip. Bottom: the ego person observes the collaborator in the first row; the collaborator's path is solid while the collaborator is observed and dashed afterwards, and the purple frustum shows the counterfactual orientation, rotated by $45^{\circ}$ about the vertical axis from the real one (blue). In both examples, the ground-truth frames follow the real trajectory and are shown for comparison only; in the bottom example, the ground truth matches the counterfactual pose in the first row.}
\label{fig:supp_cf}
\end{figure*}

Because \methodName conditions on the shared 3D scene geometry and the collaborator's camera trajectory, it can also synthesise what a collaborator would see along counterfactual trajectories that depart from the collaborator's recorded trajectory.
This capability is useful both for estimating what a collaborator would perceive from candidate positions in the workspace and for augmenting single-person recordings with diverse multi-perspective trajectories.
We illustrate two counterfactual settings without modifying the trained models, re-selecting the ego person's visual memory and re-rendering the scene coordinate maps for each counterfactual trajectory before running inference.
First, when the collaborator remained outside the ego person's view (\cref{fig:supp_cf}, top), we paired the ego person's video from a test clip with another collaborator trajectory from the test set, selected such that the collaborator's head never projected into the ego person's view and stayed at least $0.5$\,m away from the ego person.
In this example, the counterfactual trajectory stays by the refrigerator while the ego person works at the counter: in the middle row, all three backbones synthesised the opened refrigerator with hands reaching into it, and as the camera turns in the bottom row, all three backbones rendered the door while the Wan2.1-14B and AnyView versions also synthesised a person working at the counter.
Second, to illustrate both fidelity to the ground truth when the collaborator is observed and plausible synthesis once the trajectory diverges (\cref{fig:supp_cf}, bottom), we took a test clip in which the ego person observes the collaborator at the start, kept the real collaborator pose while the collaborator was observed (plus a short buffer), and then rotated the collaborator's camera by $45^{\circ}$ about the vertical axis at the same positions.
While the collaborator was observed (first row), all three backbones reproduced the counter, the sink, and the window corner of the ground truth; after the $45^{\circ}$ rotation (second and third rows), all three backbones synthesised the window side of the kitchen with the stove along the new viewing direction, whereas the ground truth continues along the original view.
Additional counterfactual video examples across all three backbones are provided on the supplementary webpage.

\subsection{Results without Scene Geometry or Visual Memory}
\label{sec:supp_missing}
Because we applied joint geometry dropout and whole-memory dropout during training (\cref{sec:supp_impl,sec:supp_memory}), the full models could be evaluated directly at inference when either the scene coordinate maps or the visual memory were unavailable.

\begin{figure*}[t]
\centering
\includegraphics[width=\textwidth]{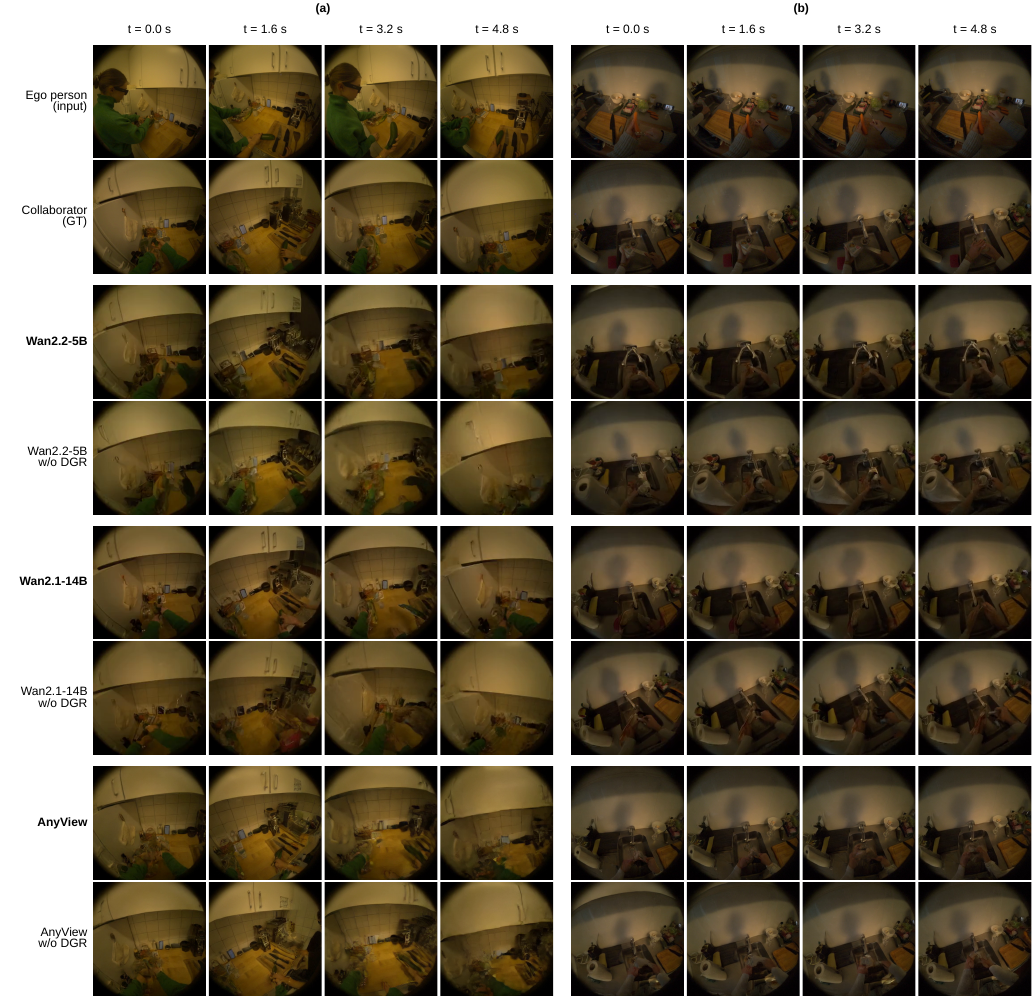}
\caption{\textbf{Qualitative comparison with and without DGR at inference.} (a) Both participants work side by side at the counter (overlapping views). (b) The ego person works at the cutting board while the collaborator washes at the sink (non-overlapping views).}
\label{fig:supp_nodgr}
\end{figure*}

\paragraph{Without scene geometry.}
\label{sec:supp_nodgr}
To evaluate \methodName when 3D scene geometry was unavailable at test time, we ran each full model on all test clips with the scene coordinate maps of the ego person, the collaborator, and the memory frames set to zero, using a single conditional forward pass without geometry guidance.
As shown in \cref{tab:supp_nodgr}, even without DGR at inference, our full models outperformed the Pl\"ucker-conditioned base models (which likewise take no 3D scene geometry) across all three backbones in PSNR ($17.28$ vs.\ $16.19$\,dB on Wan2.2-5B, $17.09$ vs.\ $16.59$\,dB on Wan2.1-14B, and $18.44$ vs.\ $18.23$\,dB on AnyView), FVD ($271.2$ vs.\ $281.0$, $485.9$ vs.\ $727.0$, and $207.1$ vs.\ $228.0$), Layout mIoU ($10.8$ vs.\ $7.2$, $10.6$ vs.\ $3.8$, and $17.9$ vs.\ $15.9$), and Hand-mIoU ($22.7$ vs.\ $20.7$, $21.7$ vs.\ $20.4$, and $27.8$ vs.\ $27.1$), showing that RGA and visual memory remained effective without dense geometric renderings.
Providing DGR at inference further improved PSNR by $1.29$, $1.80$, and $1.13$\,dB, reduced FVD to $181.2$, $164.7$, and $177.1$, and raised Layout mIoU to $20.7$, $26.1$, and $23.6$ on Wan2.2-5B, Wan2.1-14B, and AnyView, respectively.
\cref{fig:supp_nodgr} shows two qualitative examples.
In (a), where both participants work side by side at the counter, the models without DGR reconstructed the counter, tiled backsplash, and the collaborator's arms, while DGR leads to better spatial localization of the generated surface (see t = 4.8s).
In (b), where the ego person slices vegetables at the cutting board while the collaborator washes at the sink outside the ego person's current view, the models without DGR still recovered the sink and the collaborator's hands from visual memory and RGA, while DGR corrected the scale and placement of the sink basin and surrounding bowls.

\begin{table*}[t]
\centering
\small
\setlength{\tabcolsep}{3.2pt}
\resizebox{\textwidth}{!}{%
\begin{tabular}{@{}llcccccccccc@{}}
\toprule
\textbf{Backbone} & \textbf{Variant} & \textbf{PSNR}$\uparrow$ & \textbf{SSIM}$\uparrow$ & \textbf{LPIPS}$\downarrow$ & \textbf{CLIP-I}$\uparrow$ & \textbf{ViCLIP}$\uparrow$ & \textbf{FVD}$\downarrow$ & \textbf{Transl}$\downarrow$ & \textbf{Rot}$\downarrow$ & \textbf{Layout}$\uparrow$ & \textbf{Hand}$\uparrow$ \\
\midrule
\multirow{3}{*}{Wan2.2-5B} & Base model & 16.19 & 0.460 & 0.506 & 0.811 & 0.819 & 281.0 & 0.40 & 16.1 & 7.2 & 20.7 \\
 & Full model w/o DGR & 17.28 & 0.503 & 0.466 & 0.827 & 0.837 & 271.2 & 0.36 & 11.5 & 10.8 & 22.7 \\
 & Full model & \textbf{18.57} & \textbf{0.538} & \textbf{0.356} & \textbf{0.844} & \textbf{0.879} & \textbf{181.2} & \textbf{0.32} & \textbf{9.0} & \textbf{20.7} & \textbf{24.9} \\
\midrule
\multirow{3}{*}{Wan2.1-14B} & Base model & 16.59 & 0.498 & 0.525 & 0.789 & 0.710 & 727.0 & 0.34 & 19.2 & 3.8 & 20.4 \\
 & Full model w/o DGR & 17.09 & 0.506 & 0.484 & 0.819 & 0.812 & 485.9 & 0.36 & 15.3 & 10.6 & 21.7 \\
 & Full model & \textbf{18.89} & \textbf{0.551} & \textbf{0.343} & \textbf{0.853} & \textbf{0.897} & \textbf{164.7} & \textbf{0.28} & \textbf{10.1} & \textbf{26.1} & \textbf{26.9} \\
\midrule
\multirow{3}{*}{AnyView} & Base model & 18.23 & 0.513 & 0.400 & 0.872 & 0.886 & 228.0 & 0.43 & 9.3 & 15.9 & 27.1 \\
 & Full model w/o DGR & 18.44 & 0.520 & 0.385 & 0.876 & 0.892 & 207.1 & 0.44 & 9.6 & 17.9 & 27.8 \\
 & Full model & \textbf{19.57} & \textbf{0.556} & \textbf{0.306} & \textbf{0.888} & \textbf{0.916} & \textbf{177.1} & \textbf{0.41} & \textbf{8.8} & \textbf{23.6} & \textbf{30.9} \\
\bottomrule
\end{tabular}}
\caption{\textbf{Inference with and without DGR.} Comparison of the Pl\"ucker-conditioned base model, the full model evaluated with all scene coordinate maps zeroed at inference (Full model w/o DGR), and the full model with DGR across all three backbones. Bold: best per backbone.}
\label{tab:supp_nodgr}
\end{table*}

\begin{figure*}[t]
\centering
\includegraphics[width=\textwidth]{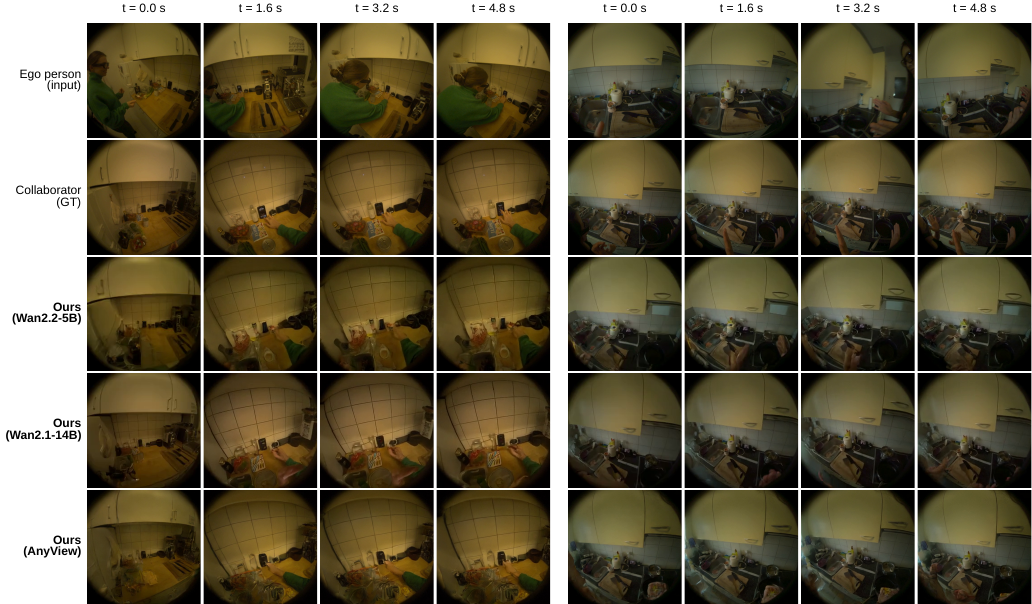}
\caption{\textbf{Qualitative results without visual memory.} Four frames sampled across two test clips at the start of a recording session.}
\label{fig:supp_nomem}
\end{figure*}

\paragraph{Without visual memory.}
\label{sec:supp_nomem}
At the start of a recording session, the ego person has no earlier frames, so the visual memory is empty.
This occurs in 29 of the 1{,}020 test clips.
Because we applied whole-memory dropout during training (\cref{sec:supp_memory}), we could evaluate our full models directly on these clips without modification.
Even without visual memory, \methodName outperformed the strongest baseline on each backbone: it achieved a PSNR of $18.10$, $18.09$, and $18.70$\,dB and an LPIPS of $0.383$, $0.388$, and $0.353$ on Wan2.2-5B, Wan2.1-14B, and AnyView, compared with $16.22$\,dB and $0.525$ for WorldWander, $17.75$\,dB and $0.452$ for EgoX, and $18.03$\,dB and $0.409$ for AnyView (all using cross-view pretrained weights).
\cref{fig:supp_nomem} shows qualitative results on two of these clips.
Without memory, \methodName still reconstructed the visible scene layout, such as the tiled wall and counter on the left and the upper cabinets, kettle, and counter on the right.

\subsection{Failure Cases}
\label{sec:supp_fail}
\begin{figure*}[t]
\centering
\includegraphics[width=\textwidth]{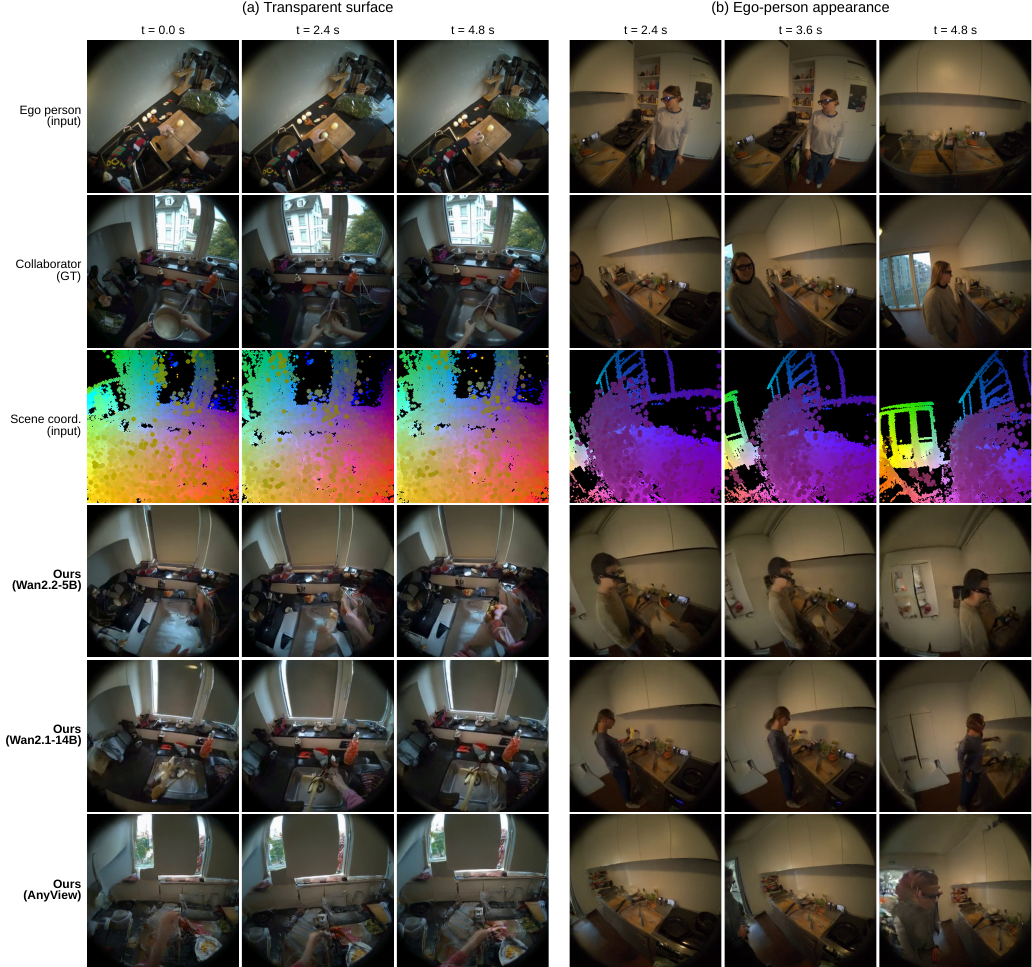}
\caption{\textbf{Failure cases.}}
\label{fig:supp_fail}
\end{figure*}

\cref{fig:supp_fail} illustrates two failure modes of \methodName: transparent surfaces that are absent from the scene point cloud (a) and unobserved ego-person appearance (b).

\paragraph{Transparent surfaces.}
In case (a), the collaborator faces a window overlooking a building outside.
Because the scene point cloud contains no points on the window glass or outside the room, the window appears as an empty region in the collaborator's scene coordinate map.
The two Wan models filled this region with a blind or a wall, while the AnyView model reconstructed the outdoor view only on the left pane.
This likely occurs because empty regions in the scene coordinate maps during training usually correspond to indoor surfaces missing from the point cloud rather than outdoor views through glass.

\paragraph{Ego-person appearance.}
In case (b), the ego person enters the collaborator's field of view at the start of a recording.
Because the ego person's egocentric camera observes only the ego person's own hands and arms and the static point cloud does not include moving people, the inputs provide no visual reference for the ego person's head or torso.
As a result, the generated person differed from the ego person in hair and facial appearance, and on Wan2.1-14B and AnyView in clothing as well.

\end{document}